\documentclass{article}
\usepackage[preprint]{neurips_2025}
\usepackage{graphicx}
\usepackage[table]{xcolor}
\usepackage{booktabs}
\usepackage{amsmath}
\usepackage{amsthm}
\usepackage{enumitem}

\usepackage{caption}
\usepackage{wrapfig}
\usepackage{tabularx}
\usepackage{booktabs}
\usepackage{listings}
\usepackage{array}
\usepackage{listings}
\usepackage{algorithm}
\usepackage{fontawesome}
\usepackage{algpseudocode}  
\usepackage{float}
\usepackage[dvipsnames]{xcolor}

\usepackage[most]{tcolorbox}
\usepackage{xcolor}

\usepackage[table]{xcolor}
\newtcolorbox{promptbox}{
    colback=green!4,
    colframe=green!45!black,
    boxrule=0.5pt,
    arc=2pt,
    left=6pt,
    right=6pt,
    top=6pt,
    bottom=6pt,
    fontupper=\ttfamily\small,
    breakable
}

\usepackage{tabularx,booktabs,array}
\newcolumntype{Y}{>{\raggedright\arraybackslash}X}
\newcolumntype{P}[1]{>{\raggedright\arraybackslash}p{#1}}

\usepackage{bbold}
\usepackage[lined,boxed,ruled,norelsize,algo2e,linesnumbered]{algorithm2e}
\usepackage{booktabs}

\usepackage{amsmath,amsfonts,bm}

\def\eqref#1{equation~\ref{#1}}

\def\1{\bm{1}}

\DeclareMathAlphabet{\mathsfit}{\encodingdefault}{\sfdefault}{m}{sl}
\SetMathAlphabet{\mathsfit}{bold}{\encodingdefault}{\sfdefault}{bx}{n}

\newtheorem{proposition}{Proposition}
\newtheorem{definition}{Definition}

\usepackage{graphicx}
\usepackage{titlesec}

\renewcommand{\paragraph}[1]{\vspace{.1em}\noindent\textbf{#1}}

\usepackage[pagebackref=true,colorlinks,citecolor=brown]{hyperref}
\usepackage{url}

\usepackage{caption}
\titlespacing\section{0pt}{3pt plus 1pt minus 1pt}{2pt plus 1pt minus 1pt}
\titlespacing\subsection{0pt}{2pt plus 1pt minus 1pt}{1pt plus 1pt minus 1pt}

\PassOptionsToPackage{numbers, compress}{natbib}

\usepackage[utf8]{inputenc} 
\usepackage[T1]{fontenc}    
\usepackage{hyperref}       
\usepackage{url}            
\usepackage{booktabs}       
\usepackage{amsfonts}       
\usepackage{nicefrac}       
\usepackage{microtype}      
\usepackage{xcolor}         

\usepackage[pagebackref=true,colorlinks,citecolor=brown]{hyperref}
\usepackage{url}

\usepackage{caption}
\titlespacing\section{0pt}{3pt plus 1pt minus 1pt}{2pt plus 1pt minus 1pt}
\titlespacing\subsection{0pt}{2pt plus 1pt minus 1pt}{1pt plus 1pt minus 1pt}

\author{%
  \textbf{Aayush Karan}\textsuperscript{1},
  \textbf{Sitan Chen}\textsuperscript{1},
  \textbf{Yilun Du}\textsuperscript{1} \\[0.6em]
  \textsuperscript{1}Harvard University \\[0.6em]
  \textbf{\href{https://aakaran.github.io/finetuning_with_sampling/}
     {\textcolor{blue!60!black}{\faGlobe\enspace{Website}}}
    \quad
  \href{https://github.com/aakaran/finetuning-with-sampling}
     {\textcolor{blue!60!black}{\faGithub\enspace{Code}}}}
}

\title{Finetuning with Sampling: \\  SFT Learns Better Than You Think}

\begin{document}

\maketitle

\begin{abstract}

Introducing new capabilities to frontier models has long been the goal of posttraining, which predominantly employs supervised finetuning (SFT) and reinforcement learning (RL) to this end. Conventional wisdom dictates that RL enables strong generalization on new tasks without losing existing capabilities, while SFT is prone to weak generalization and catastrophic forgetting. At the same time, SFT can learn from off-policy expert data, whereas RL must rely on a model's ability to find successful trajectories with repeated sampling. In our work, we seek to leverage the strength of on-policy learning while utilizing the privileged information contained in off-policy data. However, rather than modifying the learning objective to accommodate this data, we instead tailor the data distribution to better suit the learner. We introduce a Markov chain Monte Carlo (MCMC) sampling algorithm that progressively transforms off-policy traces to be more on-policy given a reference model for finetuning. Across tasks like scientific skill acquisition, mathematical reasoning, and open-ended expertise, our sampling algorithm enables SFT to rival prevailing posttraining techniques, often generalizing better and forgetting less than strong on-policy baselines. In addition, the resulting finetuned models exhibit strong distributional performance and are capable of learning beyond sharpening the base model distribution. At a higher level, our approach presents sampling as a model-native operator that \textit{shapes data for learnability}, offering broader utility as a general-purpose primitive throughout the posttraining stack.

\end{abstract}

\section{Introduction}\label{sec:intro}

Posttraining has brought about revolutionary advances in the capabilites of frontier large language models (LLMs) \citep{guo2025deepseekr1, hu2025openreasonerzero}, giving sizeable performance gains across domains like math, science, and coding \citep{hendrycks2021math, li2022alphacode, rein2024gpqa}.

Still, a key open question remains: how can we introduce \textit{fundamentally new capabilities} during posttraining without \textit{giving up existing ones}? A large body of recent literature has been dedicated to understanding the extent to which existing learning frameworks -- i.e., supervised finetuning (SFT) and reinforcement learning (RL) -- can achieve this end. Overwhelming evidence \citep{chen2025retaining, chu2025sftmemorizesrlgeneralizes, shenfeld2025rl} points towards RL as the paradigm of choice, citing superior \textit{generalization} on unseen domains that appears to leave prior abilities intact. Meanwhile, SFT suffers from weak generalization and \textit{catastrophic forgetting} \citep{kirkpatrick2017overcoming, luo2023empirical}: a phenomenon where finetuned models exhibit significant deterioration in existing capabilities.

\cite{chen2025retaining, shenfeld2025rl} attribute this performance gap to the \textit{on-policy} nature of RL: the model's own samples dictate learning updates, constraining finetuning towards distributions that do not stray far from the original starting point. SFT on the other hand is \textit{off-policy}, leaving finetuning more susceptible to drastic updates and shifts in behavior \citep{xiong2025minimalist}.

At the same time, learning strictly on-policy is also a limitation, as it relies on the model's own ability to find successful trajectories through repeated sampling. If the model cannot generate a correct rollout on a training sample, there is no collective variance in reward, resulting in no learning signal. \cite{qu2026pope} observes this deficiency in practice, identifying a sizeable regime of hard training samples that are effectively discarded during RL as they are simply too difficult to yield any trainability.

This limitation is not ideal for the purposes of introducing new capabilities, where the model is unlikely to already be competent enough to produce a successful trajectory.  Here SFT has an advantage, as it leverages trajectories constructed from off-policy \textit{privileged information} that provide rich supervision signal without necessitating an explicit reward.

In this work, rather than modifying the learning algorithm to accommodate off-policy data, we ask if we can instead \textit{shape the data distribution} to accommodate  learning. More precisely, \textit{can we algorithmically sample on-policy trajectories that remain faithful to privileged off-policy information}? To this end, we propose a sampling algorithm that progressively transforms a dataset of off-policy expert traces to be distributionally closer to a given base model distribution.

\begin{figure}[t]
  \centering
  \includegraphics[width=\linewidth]{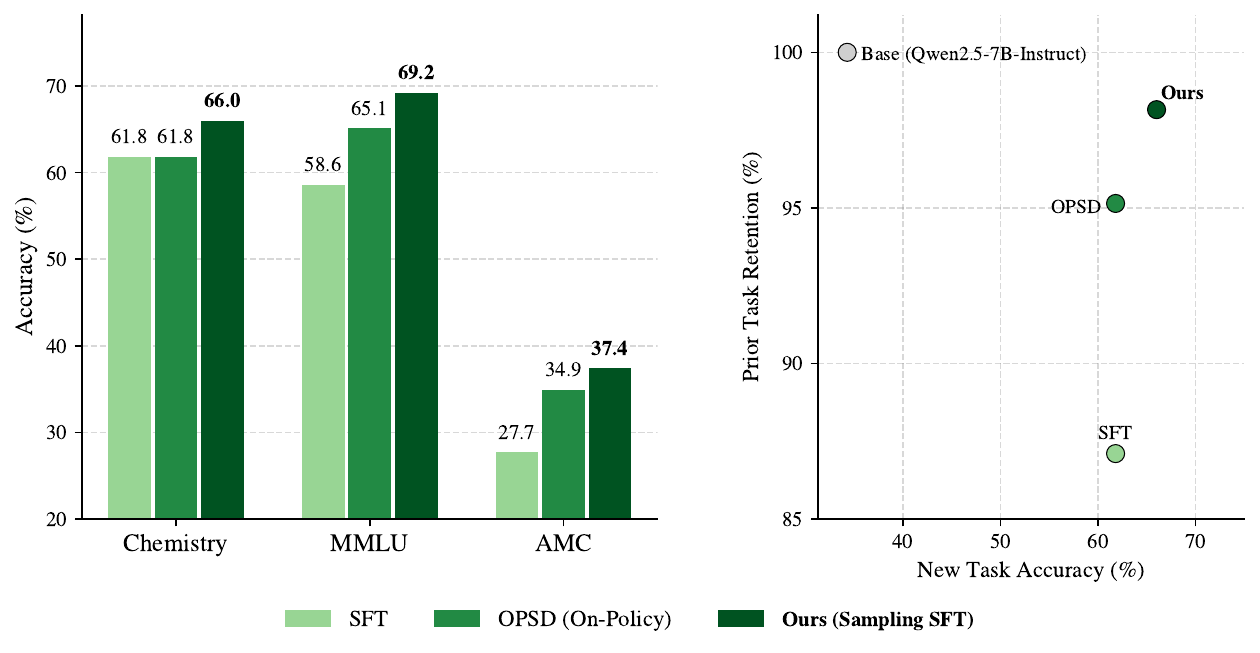}
  \captionsetup{font=small}
  \vspace{-5pt}
  \caption{\textbf{Composed with sampling, SFT can generalize better and forget less than on-policy learning.} Left: we compare SFT composed with our sampling algorithm for off-policy data against SFT on the original dataset and an on-policy learning algorithm that integrates off-policy data into self-distillation (OPSD) \citep{shenfeld2026self}. We illustrate this on the task of scientific reasoning for chemistry problems and plot performance on prior capabilities introduced during pretraining (MMLU and AMC). Right: we compare the three along two axes: new task accuracy (chemistry) and prior task retention, which measures the percent of base model performance the finetuned models are able to maintain. SFT composed with sampling performs the best on both axes.}
  \label{fig:teaser}
\end{figure}

Remarkably, this sampling step enables SFT to \textit{rival} prevailing posttraining techniques, often \textit{generalizing better and forgetting less than strong on-policy learning algorithms}.

Our contributions can be summarized as follows:
\begin{itemize}[leftmargin=*]
\vspace{-3pt}
    \item We introduce a framework that formalizes the task of transforming off-policy data into on-policy data for finetuning. Namely, given a base model and a dataset of expert trajectories, we specify a target data generation policy that preserves the information content of the original dataset while maintaining maximal proximity to the base model distribution.
\vspace{-3pt}

    \item To generate boosted samples from this policy, we introduce \textit{projection sampling}, an approximate sampling algorithm using Markov chain Monte Carlo (MCMC) techniques that iteratively refine expert trajectories according to their likelihoods with respect to the base model.   
\vspace{-3pt}

    \item We empirically demonstrate the effectiveness of our sampling procedure for SFT over a range of tasks, including  scientific skill acquisition, mathematical reasoning, and open-ended expertise. Our results show that SFT with sampling can  outperform on-policy counterparts like RL and self-distillation on new task generalization as well as prior capability retention.
\vspace{-3pt}
\end{itemize}

Our results collectively illustrate that composing sampling with finetuning is a powerful paradigm that can enable a learning algorithm as standard as SFT to perform far beyond conventional expectations.

\section{Related Works}
\label{rel_works}

\textbf{SFT vs. RL.} Many recent works have investigated the performance gap between SFT and RL in posttraining. \cite{chu2025sftmemorizesrlgeneralizes} was an early work that noticed the gap in generalization performance, asserting SFT is prone to memorization while RL can generalize beyond the training distribution. \cite{xiong2025minimalist} offers a mathematical explanation for this phenomenon, suggesting that training off-policy can result in large and unstable gradient updates that encourage overfitting and harm generalization. \cite{shenfeld2025rl} and \cite{chen2025retaining} explore the performance gap from the lens of prior capability retention, noting that SFT is highly susceptible to catastrophic forgetting \citep{kirkpatrick2017overcoming, luo2023empirical}. \cite{shenfeld2025rl} conducts extensive empirical analysis to attribute this to the off-policy nature of SFT  and even shows that the level of forgetting correlates with the KL divergence of the finetuned model with the base model. Similarly, \cite{chen2025retaining} offers a mode-seeking rationale and concludes similarly that on-policy data mitigates forgetting while off-policy data facilitates it.

Several works also attempt to bridge the gap between SFT and RL by interpolating between off-policy and on-policy learning. Both \cite{liu2025uft} and \cite{qu2026pope} construct variants of GRPO \citep{guo2025deepseekr1} that use expert data in-context to encourage generating correct rollouts for hard problems during RL, treating them as vanilla generations during training. On the SFT side, \cite{wu2025generalization} modulates the SFT objective with importance sampling weights to match the on-policy gradient, while \cite{zhu2025anchored} adds a KL divergence term to further encourage proximity to the base model. While these approaches alter the learning objective to better account for off-policy data, our approach leaves SFT intact, and instead focuses on modulating off-policy data to be more on-policy.

\paragraph{Self-Distillation.} An on-policy alternative to RL that has recently begun to gain recognition for incorporating off-policy data is \textit{self-distillation}. Self-distillation typically constructs a teacher model from a base model by conditioning on \textit{privileged information}; e.g., expert data or environment feedback, which is then distilled via a KL divergence loss. Several works \citep{shenfeld2026self, ye2026onpolicycontext, zhao2026selfdistilledreasoner} offer variations on this idea and demonstrate promising results. OPSD \citep{shenfeld2026self, zhao2026selfdistilledreasoner}, which we refer to as a baseline in this paper, offers one promising avenue for continual learning: i.e., adapting to new tasks without forgetting. 

\paragraph{MCMC Sampling with LLMs.} Prior research has also explored how to integrate MCMC methods into LLM sampling. Several works have explored using MCMC to {\it tilt} to an external reward function~\citep{zhao2024probabilisticinference, faria2024quest} or probabilistic program~\citep{lew2023sequential}. Most related to our work are \cite{karan2025reasoning}, \cite{azizi2026powersmc}, \cite{arzhantsev2026self},\nocite{karan2024unrolled} which use MCMC sampling to sample from a sharpened distribution for reasoning.  Other works expand on this framework by targeting critical tokens \citep{li2025blinkofaneyetheory}\nocite{karan2025reguidance} to make MCMC sampling more efficient \citep{zhou2026reasoning, ji2026scalable}\nocite{gollakota2025predictor}. In contrast, our approach uses sampling to transform off-policy data for subsequent finetuning.

\vspace{-5pt}
\section{Preliminaries}
\vspace{-3pt}

Let $\mathcal{X}$ be a finite vocabulary of tokens, and let $\mathcal{X}^T$ denote the set of finite sequences of tokens $x_{0:T} = (x_0, x_1, \dots, x_T)$, where $x_i \in \mathcal{X}$ for all $i$ and $T \in \mathbb{Z}_{\geq 0}$ is some nonnegative integer. For convenience, for a given $t$, let $x_{<t} = (x_0, \dots, x_{t-1})$ and $x_{>t} = (x_{t+1}, \dots, x_{T})$, with similar definitions for $x_{\leq t}$ and $x_{\geq t}$. In general, $\mathbf{x}$ refers to a token sequence $x_{0:T}$, where $T$ is implicitly given, and $\mathcal{X}^*$ denotes the set of token sequences with finite but arbitrary length.

Then an LLM defines a distribution $p$ over token sequences $\mathcal{X}^*$ by autoregressively learning the conditional token distributions $p(x_t | x_{<t})$ for all $t$, giving the \textit{joint distribution} via the identity
\begin{equation}\label{eq:joint}
    p(x_{0:T}) = \prod_{t=0}^T p(x_t | x_{<t}).
\end{equation}

To sample a sequence from $p$, we simply sample from the LLM token by token using the conditional distributions, which by (\ref{eq:joint}) directly samples from the joint distribution.

\paragraph{Supervised Finetuning.} Let $\mathcal{D} = \{(\mathbf{q}_{i}, \mathbf{x}_{i})\}_{i=1}^N$ denote a dataset of expert trajectories $\mathbf{x}_i$ in response to queries $\mathbf{q}_i$. Then SFT optimizes the cross-entropy loss objective
\begin{equation}\label{eq:sft_loss}
    \mathcal{L}_{\text{\rm SFT}} = \mathbb{E}_{(\mathbf{q}, \mathbf{x})\sim \mathcal{D}}
\left[-\log p\left(\mathbf{x} \mid \mathbf{q}\right)\right],
\end{equation}
via gradient updates on the LLM's parametrization of $p$.

\vspace{-5pt}
\section{Boosting Off-Policy Data with MCMC Sampling}
\vspace{-3pt}

\begin{figure}[h]
  \centering
  \includegraphics[width=0.75\linewidth]{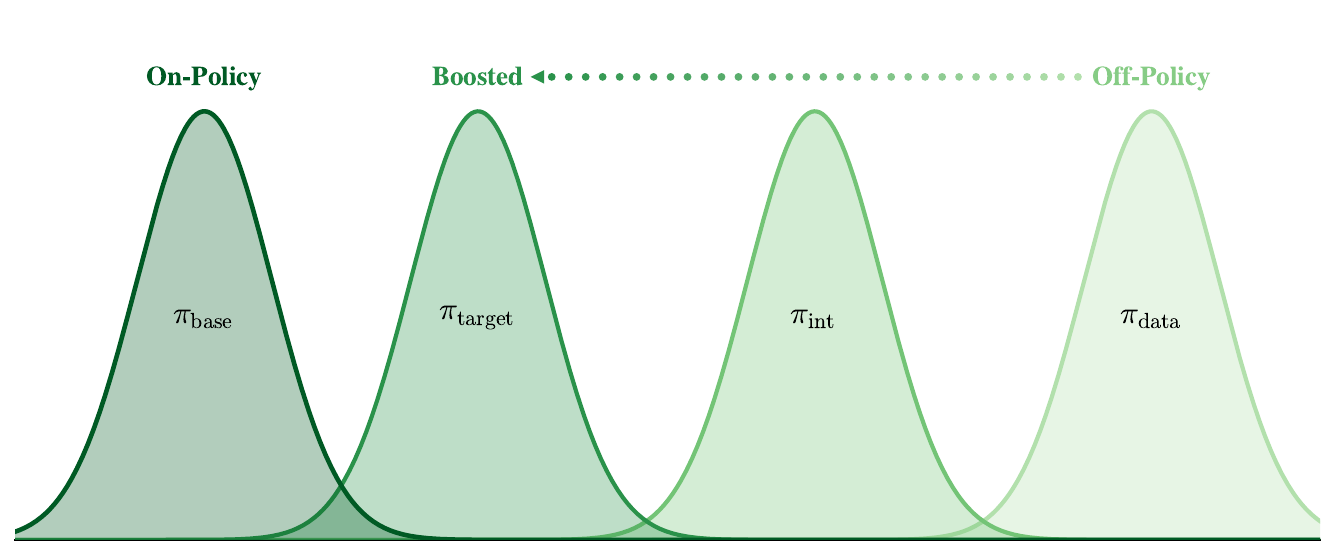}
  \captionsetup{font=small}
  \caption{\textbf{Toy schematic.} Our sampling algorithm progressively shifts the original data policy $\pi_{\rm data}$ to a target policy $\pi_{\rm target}$ that is closer to the base model distribution $\pi_{\rm base}$.}
  \label{fig:boosting}
\end{figure}

In this section, we present our sampling algorithm that maps off-policy expert data closer to the base model distribution for finetuning. Our procedure formalizes interpolating off-policy information into on-policy learning as highlighted in Section \ref{sec:intro}. To construct an informational-equivalent dataset that is easiest for the base model to internalize, we must provide a generation policy that maintains this information with minimal KL divergence. This specifies a target data generation distribution, and hence, all that remains is for us to explicitly provide an algorithm that samples from it.   

The remainder of the section is organized as follows. Section \ref{sec:min_kl} presents our formalization specifying this (unnormalized) boosted target data policy. Section \ref{sec:mh} introduces a Markov chain Monte Carlo (MCMC) algorithm that approximately samples from this policy and  ensures our sampling procedure produces generation policies that get progressively closer to the base model distribution. Finally, Section \ref{sec:prac} outlines our implementation of this algorithm for LLMs.

\subsection{Targeting the Information Projection}\label{sec:min_kl}

Suppose we are given a dataset of expert query-trajectory pairs  $\mathcal{D} = \{(\mathbf{q}_{i}, \mathbf{x}_{i})\}_{i=1}^N$. For notational convenience, we fix a query $\mathbf{q}_i$ for some $i$ for the remainder of the section and assume all distributions are implicitly conditioned on $\mathbf{q}_{i}$. We can easily lift this assumption by stitching together the conditionals across $1 \leq i \leq N$.

For a given equivalence relation on trajectories, let $\mathcal{C}$ define the set of all trajectories $\mathbf{x}\in \mathcal{X}^*$ that are equivalent to the associated expert trajectory $\mathbf{x}_{i}$. For theoretical purposes, we assume the existence of such an equivalence relation a priori; however, in practice, $\mathcal{C}$ refers to a task-dependent notion of ''correctness'' or ''semantic equivalence''. For example, for math/science reasoning, a semantically equivalent trajectory $\mathbf{x}$ refers to any trajectory leveraging information from expert trace $\mathbf{x}_{i}$ that results in a correct final answer. For fact learning, a semantic equivalent $\mathbf{x}$ must contain the same factual content as $\mathbf{x}_{i}$, as judged by an automated (LLM) grader.

 \begin{definition}
    {\rm Let $\mathcal{P}_{\mathcal{C}}$ denote the space of all distributions over token sequences $\mathcal{X}^*$ that have conditional support on $\mathcal{C}$: that is, for a policy $\pi \in \mathcal{P}_{\mathcal{C}}$, we have $\pi(\mathbf{x}) = 0$ if $ \mathbf{x} \notin \mathcal{C}$. In other words, informally, $\mathcal{P}_{\mathcal{C}}$ is the set of all distributions that output only trajectories that are equivalent to the expert traces. }
\end{definition}

Our setup naturally admits a well-specified formalism for boosting off-policy data. The dataset of expert trajectories itself is drawn from some arbitrary policy $q \in \mathcal{P}_{\mathcal{C}}$, implying the SFT objective (\ref{eq:sft_loss}) minimizes the cross-entropy $\mathbb{E}_q[- \log p]$. However, for arbitary $q$, this objective can be arbitarily off-policy.  Given that we want to preserve the information signal of $\mathcal{D}$  with trajectories that are as on-policy as possible, this is equivalent to finding the distribution in $\mathcal{P}_{\mathcal{C}}$ closest to the base model distribution $p$. 

\begin{definition}
{\rm The \textit{information projection} ~\citep{csiszar1975divergence} of $p$ onto
$\mathcal{P}_{\mathcal{C}}$ is defined as
\begin{equation}\label{eq:opt}
    p_{\mathcal{C}}
    :=
    \arg\min_{\pi \in \mathcal{P}_{\mathcal{C}}}
    \mathrm{KL}\!\left(
        \pi
        \,\middle\|\,
        p
    \right).
\end{equation}}
\end{definition}

\begin{proposition}
The distribution that uniquely minimizes the objective (\ref{eq:opt})
is the restriction of the base model onto equivalent trajectories:
\[
    p_{\mathcal{C}}(\mathbf{x})
    =
    \frac{
        p(\mathbf{x}) \cdot 
        \mathbf{1}\bigl(\mathbf{x} \in \mathcal{C}\bigr)
    }{
        \sum_{\mathbf{x}' \in \mathcal{X}^*}
    p(\mathbf{x}') \cdot 
    \mathbf{1}\bigl(\mathbf{x}' \in \mathcal{C}\bigr)
    },
\]
where $\mathbf{1}\bigl(\mathbf{x} \in \mathcal{C}\bigr)$  takes the value $1$ if $\mathbf{x} \in \mathcal{C}$ and $0$ otherwise.
\end{proposition}

\begin{proof}
Note that for any $\pi \in \mathcal{P}_{\mathcal{C}}$ we have that \[
\begin{aligned}
    \mathrm{KL}\!\left(
        \pi
        \,\middle\|\,
        p
    \right)
    &=
    \sum_{\mathbf{x}\in \mathcal{X}^*}
    \pi(\mathbf{x})
    \log
    \frac{
        \pi(\mathbf{x})
    }{
        p(\mathbf{x})
    }
    =
    \sum_{\mathbf{x}\in \mathcal{C}}
    \pi(\mathbf{x})
    \log
    \frac{
        \pi(\mathbf{x})
    }{
        p(\mathbf{x})
    },
\end{aligned}
\]
since $\pi(\mathbf{x}) = 0$ if $\mathbf{x} \notin \mathcal{C}$. Now $\pi$ defines a distribution over $\mathbf{x} \in \mathcal{C}$, but $p$ is not normalized under the restriction to correct trajectories. However, we can rewrite $p(\mathbf{x}) = Z \cdot p_{\mathcal{C}}(\mathbf{x})$ for $\mathbf{x} \in \mathcal{C}$, where $Z = \sum_{\mathbf{x}'\in \mathcal{C}} p(\mathbf{x}')$, so $\mathrm{KL}\!\left(
        \pi
        \,\middle\|\,
        p
    \right)$ simplifies to 
 \begin{equation}\label{eq:equiv}
    \sum_{\mathbf{x}\in \mathcal{C}}
    \pi(\mathbf{x})
    \log
    \frac{
        \pi(\mathbf{x})
    }{
        Z \cdot p_{\mathcal{C}}(\mathbf{x})
    }
    =-\log Z  + \sum_{\mathbf{x}\in \mathcal{C}}
    \pi(\mathbf{x})
    \log
    \frac{
        \pi(\mathbf{x})
    }{
        p_{\mathcal{C}}(\mathbf{x})
    } = \mathrm{KL}\!\left(
        \pi
        \,\middle\|\,
        p_{\mathcal{C}}
    \right) -\log Z,
\end{equation}
since $\sum_{\mathbf{x}\in \mathcal{C}} \pi(\mathbf{x}) = 1$ and both $\pi$ and $p_{\mathcal{C}}$ define a distribution over $\mathcal{C}$. Then since $\log Z$ is a constant independent of $\pi$, it follows that the expression is uniquely minimized at $\pi = p_{\mathcal{C}}$, as desired.
\end{proof}

The unnormalized projection $p_{\mathcal{C}}(\mathbf{x}) \propto p(\mathbf{x}) \cdot \mathbf{1}\bigl(\mathbf{x} \in \mathcal{C}\bigr)$ is thus our target distribution, as it is the closest distribution to the base model that also maintains the information content of the off-policy expert traces. Of course, if the base model $p$ can reliably generate expert trajectories itself, the easiest way to sample from $p_{\mathcal{C}}$ is to simply perform rejection sampling, which already lends itself to a rudimentary form of on-policy RL~\citep{xiong2025minimalist}.

However, we are interested in the regime where rejection sampling is infeasible, and the expert trajectories are difficult to attain via the base model alone. Then our task is now the following: given access to an arbitrary initial expert policy $q$ via dataset trajectories, can we algorithmically evolve samples from $q$ into informational-equivalent but more on-policy samples from $p_{\mathcal{C}}$?

\subsection{The Metropolis-Hastings Algorithm}\label{sec:mh}

We indeed can, by appealing to Markov Chain Monte Carlo (MCMC) approximate sampling techniques. Specifically, we will use a Metropolis-Hastings (MH) sampler~\citep{metropolis1953equation,hastings1970monte}, which constructs a Markov chain of sample sequences $(\mathbf{x}^0, \mathbf{x}^1, \dots, \mathbf{x}^n)$ using a \textit{proposal distribution} $\kappa(\mathbf{x}|\mathbf{x}^i)$ to select the next candidate $\mathbf{x}^{i+1}$. With probability 
\begin{equation}\label{eq:accept}
    A(\mathbf{x}, \mathbf{x}^i) = \text{min} \left\lbrace 1, \frac{p_{\mathcal{C}}(\mathbf{x}) \cdot \kappa(\mathbf{x}^i | \mathbf{x})}{p_{\mathcal{C}}(\mathbf{x}^i) \cdot \kappa(\mathbf{x}|\mathbf{x}^i)}\right\rbrace,
\end{equation}
 
candidate $\mathbf{x}$ is accepted as $\mathbf{x}^{i+1}$; otherwise, MH sets $\mathbf{x}^{i+1} = \mathbf{x}^{i}$.  It is a classic fact that as $n \to \infty$, this process converges to sampling from the \textit{target distribution} $p_{\mathcal{C}}$, provided that the Markov chain satisfies the following properties:

 \begin{definition}
    {\rm The proposal distribution $\kappa$ is \textit{irreducible} if under the induced Markov chain, for any states $\mathbf{x},\mathbf{x}'$ with nonzero mass under the target distribution $p_{\mathcal{C}}$, the probability of transitioning to $\mathbf{x}'$ starting from $\mathbf{x}$ after some number of steps is nonzero. The proposal is \textit{aperiodic} if the induced chain of samples does not return to the same sample after a fixed interval number of steps.}
\end{definition}

Since we are given a dataset of expert traces, we would like to initialize the Markov chain with such a trace and restrict our attention to proposal distributions $\kappa_{\mathcal{C}}$ that ensure that $\mathbf{x}^i \in \mathcal{C}$ for all $i$. We refer to such proposal distributions as \emph{information-preserving}.

Although this might seem to violate the irreducibility criterion, notice we can rewrite (\ref{eq:accept}) as the following: 
\begin{equation}\label{eq:new_accept}
    A(\mathbf{x}, \mathbf{x}^i) = \text{min} \left\lbrace 1, \frac{p(\mathbf{x}) \cdot \kappa(\mathbf{x}^i | \mathbf{x}) \cdot 
    \mathbf{1}\bigl(\mathbf{x} \in \mathcal{C}\bigr)}{p(\mathbf{x}^i) \cdot \kappa(\mathbf{x}|\mathbf{x}^i) \cdot 
    \mathbf{1}\bigl(\mathbf{x}^i \in \mathcal{C}\bigr)}\right\rbrace = \text{min} \left\lbrace 1, \frac{p(\mathbf{x}) \cdot \kappa_{\mathcal{C}}(\mathbf{x}^i | \mathbf{x})}{p(\mathbf{x}^i) \cdot \kappa_{\mathcal{C}}(\mathbf{x}|\mathbf{x}^i)}\right\rbrace.
\end{equation}
Assuming current candidate $\mathbf{x}^i \in \mathcal{C}$, if the new candidate $\mathbf{x} \notin \mathcal{C}$, the acceptance ratio defaults to $0$. In particular: 

\begin{proposition}
Running Metropolis-Hastings with irreducible proposals $\kappa$ and target distribution $p_{\mathcal{C}}$ is equivalent to running MH with information-preserving proposals $\kappa_{\mathcal{C}}$ and target distribution $p$.
\end{proposition}

This gives us a convenient recipe for boosting an off-policy dataset: initialize at the expert trajectory, mutate the candidates with some information-preserving proposal distribution $\kappa_{\mathcal{C}}$, and use (\ref{eq:new_accept}) to accept or reject candidates. In fact, not only does this process eventually converge to $p_{\mathcal{C}}$, but it also guarantees a progressive decrease in the policy gap between the data generation policy and the base model.

\begin{proposition}\label{prop:scale}
    For all $k \geq 0$, let $\pi_k$ denote the distribution over candidates obtained after $k$ MCMC steps applied to initial expert policy $q$. Then we have
    \begin{equation}
        \mathrm{KL}\!\left(
        \pi_{k+1}
        \,\middle\|\,
        p
    \right) \leq \mathrm{KL}\!\left(
        \pi_{k}
        \,\middle\|\,
        p
    \right).
    \end{equation}
\end{proposition}

\begin{proof}
    Let $\Psi$ denote the MCMC transition kernel; i.e., the transformation induced by applying one step of our Metropolis-Hastings procedure. $\Psi$ has stationary distribution $p_{\mathcal{C}}$, so $\Psi(p_{\mathcal{C}}) = p_{\mathcal{C}}$. Since $\Psi$ is Markovian, it follows from the data processing inequality that 
    \begin{equation}
        \mathrm{KL}\!\left(
        \pi_{k+1}
        \,\middle\|\,
        p_{\mathcal{C}}
    \right) = \mathrm{KL}\!\left(
        \Psi(\pi_k) 
        \,\middle\|\,
        \Psi(p_{\mathcal{C}})
    \right) \leq \mathrm{KL}\!\left(
        \pi_k
        \,\middle\|\,
        p_{\mathcal{C}}
    \right).
    \end{equation}
    Now, for any $\pi$, recall from (\ref{eq:equiv}) that $\mathrm{KL}\!\left(
        \pi
        \,\middle\|\,
        p_{\mathcal{C}}
    \right) = \mathrm{KL}\!\left(
        \pi
        \,\middle\|\,
        p
    \right) + \log Z$
for some fixed constant $Z$, so it follows that $\mathrm{KL}\!\left(
        \pi_{k+1}
        \,\middle\|\,
        p
    \right)  \leq \mathrm{KL}\!\left(
        \pi_k
        \,\middle\|\,
        p
    \right)$, as claimed.
\end{proof}

In other words, the Metropolis-Hastings procedure progressively shifts the original expert policy to informational-equivalent policies that get progressively closer to the base model distribution.   

\subsection{Projection Sampling for LLMs}\label{sec:prac}
As detailed in \cite{karan2025reasoning}, a direct implementation of Metropolis-Hastings for LLMs is infeasible as it requires regenerating full-length token sequences with repeated LLM inference calls. This infeasibility is further compounded by  \textit{slow mixing}, where convergence can require exponentially many MCMC steps due to  \textit{high dimensional} sample spaces or poor choice of proposals and initializations.

To help avoid these issues, we adapt the implementation of Metropolis-Hastings for power sampling in \cite{karan2025reasoning}, which we refer the reader to for further details. In short, we define restriction distributions of the target projection $p_{\mathcal{C}}$ over sequences of increasing length in units of block size $B$. We use Metropolis-Hastings to sample within each distribution, providing a strong initialization to begin sampling from the next one. The full details are presented in Algorithm \ref{alg:samp}.

\begin{figure}[t]
\begin{minipage}{\linewidth}
\begin{algorithm2e}[H]
\DontPrintSemicolon
\caption{Projection Sampling with Autoregressive Models}
\label{alg:samp}
\SetKwInOut{Input}{Input}
\SetKwInOut{Output}{Output}
\SetKwInOut{Params}{Hyperparams}

\Input{base $p$; proposal $\kappa_{\mathcal{C}}$; expert trajectory $\tau$; length $T$}
\BlankLine
\Params{block size $B$; MCMC steps $N_{\mathrm{MCMC}}$}
\BlankLine
\Output{$(x_0,\dots,x_T) \sim p_{\mathcal{C}}$}

\BlankLine
\textbf{Notation:} Define the unnormalized intermediate target
\[{\pi}_k(x_{0:{kB}}) \;\propto\; p_{\mathcal{C}}(x_{0:kB}).
\]

\BlankLine

\For{$k \gets 0$ \KwTo $\lceil \frac{T}{B}\rceil-1$}{
  
  Given prefix $x_{0:kB}$, we wish to sample from $\pi_{k+1}$. Construct initialization ${\mathbf{x}}^{0}$ by extending autoregressively with $\kappa_{\mathcal{C}}$:
  \[
  x^{(0)}_t \sim \kappa_{\mathcal{C}}\big(x_t \mid x_{<t}, \tau\big), 
    \qquad\text{for  } kB+1 \leq t \leq (k+1)B.
  \]
  Set the current state $\mathbf{x} \gets \mathbf{x}^{0}$. \;

  \BlankLine
  \For{$n \gets 1$ \KwTo $N_{\mathrm{MCMC}}$}{
    Sample an index $m \in \{1,\dots, (k+1)B\}$  uniformly. \;
    \BlankLine
    Construct proposal sequence $\mathbf{x}'$ with prefix $x_{0:m-1}$ and resampled completion:
    \[
    x'_{t} \sim \kappa_{\mathcal{C}}\big(x_t \mid x_{<t}, \tau\big), 
    \qquad\text{for } m \leq t \leq (k+1)B.
    \]

    Compute acceptance ratio (\ref{eq:accept})
    \[
      A(\mathbf{x'}, \mathbf{x}) \;\gets\; \min\Bigg\{1,\ 
      \frac{p(\mathbf{x'})}{p(\mathbf{x})}
      \cdot 
      \frac{\kappa_{\mathcal{C}}(\mathbf{x}\mid \mathbf{x'}, \tau)}{\kappa_{\mathcal{C}}(\mathbf{x'}\mid \mathbf{x}, \mathbf{\tau})}
      \Bigg\}.
    \]
    Draw $u \sim \mathrm{Uniform}(0,1)$; \;
    \lIf{$u \le A(\mathbf{x'}, \mathbf{x})$}{\textbf{accept} and set $\mathbf{x} \gets \mathbf{x'}$}
  }
  Set $x_{0:(k+1)B} \gets \mathbf{x}$ to fix the new prefix sequence for the next stage. \;
}
\Return{$x_{0:T}$} \;
\end{algorithm2e}
\end{minipage}
\end{figure}

 It remains to define an appropriate proposal distribution, as this crucial object facilitates information-preservation and is the only component of Algorithm \ref{alg:samp} that interacts with the expert trajectory. 
 
 To craft our choice of $\kappa_{\mathcal{C}}$, we use a  trick also utilized in \cite{qu2026pope, shenfeld2026self} which leverages the strong in-context instruction-following capabilities of pretrained LLMs. Suppose we are given candidate $\mathbf{x} = (x_0, \cdots x_L)$. Our proposal distribution uniformly at random selects an index $t \in [1, L]$ and creates a partial trace by truncating the suffix after $t$: $\tilde{\mathbf{x}} = (x_0, \cdots, x_t)$. Then, we pass this partial trace along with the expert trajectory $\tau$ in-context to a simple prompt template similar to ones used in \cite{qu2026pope}:
\begin{promptbox}
You are given a question, an expert solution, and an initial, partial response. Examine the solution, identifying all information provided in the reasoning process. 

\bigskip

Question: <QUERY>

\bigskip

This is an expert solution to the query: <TRAJECTORY>

\bigskip

Here is a partial response: <PARTIAL TRACE>

\bigskip

Starting with the partial response, continue in your own words, including the thinking process. Ensure your response is logically consistent with the expert solution and leads to a complete and correct final answer.
\end{promptbox}

Then our proposal $\kappa_{\mathcal{C}}$ is simply the  distribution defined by conditioning our base model $p$ on this prompt. In practice, we find that this prompt can sufficiently generate correct trajectories from expert demonstrations, satisfying the constraint of information-preservation. 

With this final piece specified, to construct our boosted dataset for finetuning, we simply apply Algorithm \ref{alg:samp} to each query-trajectory pair in our expert dataset $\mathcal{D}$. For examples of boosted traces across different models and datasets, see Appendix \ref{apx:examp}.

Since our Metropolis-Hastings procedure targets sampling from the information projection given information constraints $\mathcal{C}$, we refer to our sampling algorithm as \textit{projection sampling}. 

Unlike in \cite{karan2025reasoning}, where the sampling algorithm incurs an inference-time cost, projection sampling only incurs a \textit{one-time cost}, as it is conducted on the training dataset prior to finetuning. To quantify this cost, we can estimate the average number of tokens expended by Algorithm \ref{alg:samp}. For each trajectory $\tau \in \mathcal{D}$, each MCMC step resamples an average of $\frac{kB}{2}$ tokens, carried out $N_{\text{MCMC}}$ times. Summing, we have 
\begin{equation}\label{eq:tok}\mathbb{E}_{\text{cost}} = |\mathcal{D}|\cdot N_{\text{MCMC}}\sum_{k=1}^{\lceil T/B \rceil} \frac{kB}{2} \approx |\mathcal{D}| \cdot \frac{N_{\text{MCMC}}T^2}{4B}.\end{equation}

We can lower the cost by increasing the block size $B$, but in general, since projection sampling is a fixed cost, we prefer smaller block sizes that provide higher resolution and less approximation error.

\section{Experiments}
\label{headings}

We now empirically demonstrate that the trajectories generated by our sampling algorithm enable SFT to break its weak characterization, frequently outperforming existing on-policy learning algorithms for posttraining.

\subsection{Experimental Setup}

\paragraph{Tasks.} We select a set of learning tasks that are characteristic of the range of capabilities posttraining seeks to introduce, including novel skill acquisition, reasoning, and open-ended expertise. In the following discussion, we present three tasks. Each learning task is accompanied by expert trajectories which are either inherent to the dataset or generated by GPT-5.
\vspace{-8pt}

\begin{itemize}[leftmargin=*]
    
    \item \textbf{Chemistry:} We examine whether a pretrained LLM can pick up a new set of skills that it is not explicitly finetuned on beforehand. Following \cite{shenfeld2026self}, we take \textit{scientific reasoning} as the test domain. In this setting, we use Chemistry L-3 subset of SciKnowEval~\citep{feng2024sciknoweval} that includes skills like reaction prediction, molar weight calculation, and chemical equation balancing. Moreover, some questions follow a multiple choice format, while others (balancing) are freeform, representing a mixed dataset with a harder-to-specify reward for on-policy RL. There are 2400 problems total, with 1800 training problems and 600 test problems. Expert trajectories are generated by GPT-5. 

    \item \textbf{Math:} We investigate whether a pretrained LLM with some general mathematical ability can improve performance on a set of \textit{hard math reasoning problems} relative to the base model. In this setting, we use the MATH~\citep{hendrycks2021math} dataset restricted to the hardest problem classes: Levels 3, 4, and 5. There are 9254 problems total, with 8230 training problems and 1024 test problems that are accompanied by existing expert solutions. 

    \item \textbf{Medical:} We test whether our approach is effective in more open-ended domains that are not as immediately verifiable as the prior two. As in \cite{shenfeld2026self}, we look at \textit{medical question answering}. We use the HuatuoGPT-o1 SFT dataset \citep{chen2024huatuogpt}, which provides 19704 questions requiring clinical reasoning skills  for diagnoses, treatments, and general medical knowledge, and each question has a provided expert response. For evaluation, we select 1000 questions uniformly at random from the HuatuoGPT-o1 RL dataset that are never seen during training.

\end{itemize}
\vspace{-5pt}

\paragraph{Models.} For the given posttraining tasks, we select base models across varying model families and sizes that are not already proficient or finetuned on the task domains. For chemistry, we use Qwen2.5-7B-Instruct \citep{yang2024qwen25} and Olmo-3-7B-Instruct \citep{olmo2025olmo3}, which achieve test set accuracies $34.3\%$ and $32.8\%$ respectively. For math, we use Qwen2.5-3B, which achieves test accuracy $31.5\%$. Finally, for medical, we use Qwen2.5-7B-Instruct, which achieves test accuracy $35.3\%$.

\paragraph{Evaluation.} We include several benchmarks for each task that holistically measure both \textit{task-specific generalization} and \textit{retention of prior capabilities}. All benchmarks are scored based on {single-shot accuracy}.
\vspace{-8pt}

\begin{itemize}[leftmargin=*]
 \item \textbf{Chemistry:} We measure generalization capabilities in domain with the test set for SciKnowEval. To measure prior capabilities, we evaluate on math benchmarks AMC, MATH500, and GSM8K ~\citep{cobbe2021training} alongside MMLU ~\citep{hendrycks2020measuring} and GPQA ~\citep{rein2024gpqa}.
 
    \item \textbf{Math:} To measure generalization capabilities, we track performance on the test set MATH(3,4,5) as well as out-of-distribution generalization on AMC, MATH500, and GSM8K. To track prior capabilities, we evaluate on MMLU, Chemistry, and GPQA. 

     \item \textbf{Medical:} Since medical reasoning is a more open-ended domain, we use an automated LLM grader (GPT-5-mini) to evaluate correctness on the test set (see Appendix \ref{apx:prompts} for the exact grading prompt). For capability retention, we evaluate AMC, MATH500, GSM8K, MMLU, and GPQA.    
\end{itemize}

\textbf{Baselines.} For each task, we benchmark against a wide variety of posttraining algorithms ranging from vanilla SFT to on-policy learning with privileged information via RL and self-distillation. 
\vspace{-7pt}

\begin{itemize}[leftmargin=*]
    
    \item \textbf{Chemistry:} 
    We compare our method on all evaluations for this task to the base models, SFT with the expert dataset trajectories, and a variant of SFT using our sampling proposal prompt to rewrite each expert trajectory; i.e., ``$0^{\text{\rm th}}$ order MCMC''. In addition, we include on-policy self-distillation (OPSD) \citep{shenfeld2026self, zhao2026selfdistilledreasoner}, which distills from the base model conditioned on privileged information.

    \item \textbf{Math:} We use the baselines used in chemistry alongside two RL baselines, since the rollouts are easily verifiable. We include GRPO \citep{shao2024deepseekmath} as well as a stronger variant UFT \citep{liu2025uft}, which uses the off-policy expert traces as privileged information during rollout generation.

    \item \textbf{Medical:} Same baselines as chemistry.

\end{itemize}

\paragraph{Sampling.} For all datasets, we use block number $B = 32$ with maximum sequence length $T = 1856$ over $N_{\text{\rm MCMC}} = 10$ MCMC steps.

\paragraph{Training.} For all tasks, all SFT training runs are tuned with the following hyperparameters across the corresponding ranges: epoch number: $\{1, 2\}$, learning rate: $\{$5e-5, 1e-5, 5e-6$\}$, and batch size: $\{16, 32, 64\}$. For the medical task, we extend the epoch range to $\{1, 2, 4, 6\}$ to accommodate a larger dataset. The optimizer is AdamW with a cosine scheduler and gradient clipping for norms greater than $1$. The on-policy self-distillation runs are tuned with hyperparameters from \citep{shenfeld2026self} while the RL baselines use the default hyperparameters from \citep{liu2025uft}. All models are trained using H100s and H200s.

\subsection{Main Results}
\vspace{-2pt}

\begin{table*}[t]
\centering

\resizebox{\textwidth}{!}{
\rowcolors{3}{gray!15}{white}
\begin{tabular}{ll|c|cccccc}
\toprule
\textbf{Model} & \textbf{Algorithm}
& \multicolumn{1}{c|}{\textbf{New Task Domain}}
& \multicolumn{6}{c}{\textbf{Prior Capabilities}} \\
\cmidrule(lr){3-3} \cmidrule(lr){4-9}
& & Chemistry
& MMLU & GPQA & AMC & MATH500 & GSM8K & Avg. \\
\midrule

& Base
& \textcolor[HTML]{707070}{0.343}
& \textcolor[HTML]{707070}{0.687}
& \textcolor[HTML]{707070}{0.253}
& \textcolor[HTML]{707070}{0.422}
& \textcolor[HTML]{707070}{0.734}
& \textcolor[HTML]{707070}{0.918}
& \textcolor[HTML]{707070}{0.597} \\

& SFT
& \textcolor[HTML]{83CB82}{0.618}
& \textcolor[HTML]{98D594}{0.586}
& \textcolor[HTML]{65BD6F}{0.242}
& \textcolor[HTML]{98D594}{0.277}
& \textcolor[HTML]{98D594}{0.622}
& \textcolor[HTML]{98D594}{0.871}
& \textcolor[HTML]{98D594}{0.520} \\

Qwen2.5-7B-Instruct
& Rewrite SFT
& \textcolor[HTML]{98D594}{0.613}
& \textcolor[HTML]{339C52}{0.640}
& \textcolor[HTML]{238B45}{0.273}
& \textcolor[HTML]{4AAF61}{0.313}
& \textcolor[HTML]{2B934B}{0.676}
& \textcolor[HTML]{309950}{0.887}
& \textcolor[HTML]{2B934B}{0.558} \\

& OPSD (on-policy)
& \textcolor[HTML]{83CB82}{0.618}
& \textcolor[HTML]{268E47}{0.651}
& \textcolor[HTML]{98D594}{0.222}
& \textcolor[HTML]{137D39}{0.349}
& \textbf{\textcolor[HTML]{005321}{0.716}}
& \textbf{\textcolor[HTML]{005321}{0.901}}
& \textcolor[HTML]{157F3B}{0.568} \\

& \textbf{Sampling SFT (Ours)}
& \textbf{\textcolor[HTML]{005321}{0.660}}
& \textbf{\textcolor[HTML]{005321}{0.692}}
& \textbf{\textcolor[HTML]{005321}{0.303}}
& \textbf{\textcolor[HTML]{005321}{0.374}}
& \textcolor[HTML]{339C52}{0.670}
& \textcolor[HTML]{1E8741}{0.891}
& \textbf{\textcolor[HTML]{005321}{0.586}} \\

\bottomrule
\end{tabular}
}

\vspace{1.2em}

\resizebox{\textwidth}{!}{
\rowcolors{3}{gray!15}{white}
\begin{tabular}{ll|ccccc|cccc}
\toprule
\textbf{Model} & \textbf{Algorithm}
& \multicolumn{5}{c|}{\textbf{New Task Domain}}
& \multicolumn{4}{c}{\textbf{Prior Capabilities}} \\
\cmidrule(lr){3-7} \cmidrule(lr){8-11}
& &
MATH(3,4,5) & AMC & MATH500 & GSM8K & Avg.
& Chemistry & MMLU & GPQA & Avg. \\
\midrule

& Base
& \textcolor[HTML]{707070}{0.315}
& \textcolor[HTML]{707070}{0.133}
& \textcolor[HTML]{707070}{0.245}
& \textcolor[HTML]{707070}{0.579}
& \textcolor[HTML]{707070}{0.318}
& \textcolor[HTML]{707070}{0.283}
& \textcolor[HTML]{707070}{0.651}
& \textcolor[HTML]{707070}{0.333}
& \textcolor[HTML]{707070}{0.422} \\

& SFT
& \textcolor[HTML]{98D594}{0.243}
& \textcolor[HTML]{88CE87}{0.100}
& \textcolor[HTML]{98D594}{0.168}
& \textcolor[HTML]{98D594}{0.455}
& \textcolor[HTML]{98D594}{0.242}
& \textcolor[HTML]{98D594}{0.222}
& \textcolor[HTML]{98D594}{0.648}
& \textcolor[HTML]{98D594}{0.298}
& \textcolor[HTML]{98D594}{0.389} \\

& OPSD (on-policy)
& \textcolor[HTML]{88CE87}{0.267}
& \textcolor[HTML]{98D594}{0.084}
& \textcolor[HTML]{50B264}{0.332}
& \textcolor[HTML]{73C476}{0.524}
& \textcolor[HTML]{73C476}{0.302}
& \textcolor[HTML]{55B567}{0.242}
& \textcolor[HTML]{097532}{0.652}
& \textcolor[HTML]{60BA6C}{0.313}
& \textcolor[HTML]{4BB062}{0.404} \\

Qwen2.5-3B
& GRPO (on-policy)
& \textcolor[HTML]{17813D}{0.457}
& \textcolor[HTML]{0B7734}{0.249}
& \textcolor[HTML]{5AB769}{0.313}
& \textcolor[HTML]{005F26}{0.808}
& \textcolor[HTML]{1F8742}{0.457}
& \textcolor[HTML]{00682A}{0.278}
& \textcolor[HTML]{097532}{0.652}
& \textcolor[HTML]{60BA6C}{0.313}
& \textcolor[HTML]{268E47}{0.414} \\

& UFT (on-policy)
& \textcolor[HTML]{117B38}{0.470}
& \textbf{\textcolor[HTML]{005321}{0.293}}
& \textcolor[HTML]{62BB6D}{0.297}
& \textcolor[HTML]{0D7836}{0.746}
& \textcolor[HTML]{218944}{0.452}
& \textcolor[HTML]{005A24}{0.283}
& \textbf{\textcolor[HTML]{005321}{0.653}}
& \textcolor[HTML]{309950}{0.328}
& \textcolor[HTML]{0C7735}{0.421} \\

& \textbf{Sampling SFT (Ours)}
& \textbf{\textcolor[HTML]{05712F}{0.495}}
& \textcolor[HTML]{006227}{0.277}
& \textbf{\textcolor[HTML]{016E2D}{0.582}}
& \textcolor[HTML]{006B2B}{0.782}
& \textbf{\textcolor[HTML]{006729}{0.534}}
& \textcolor[HTML]{19833E}{0.266}
& \textcolor[HTML]{289049}{0.651}
& \textbf{\textcolor[HTML]{097532}{0.343}}
& \textcolor[HTML]{117B38}{0.420} \\

& \textbf{Sampling SFT + RL (Ours)}
& \textbf{\textcolor[HTML]{005321}{0.545}}
& \textcolor[HTML]{117B38}{0.241}
& \textbf{\textcolor[HTML]{005321}{0.652}}
& \textbf{\textcolor[HTML]{005321}{0.830}}
& \textbf{\textcolor[HTML]{005321}{0.567}}
& \textbf{\textcolor[HTML]{005321}{0.285}}
& \textcolor[HTML]{097532}{0.652}
& \textbf{\textcolor[HTML]{005321}{0.354}}
& \textbf{\textcolor[HTML]{005321}{0.430}} \\

\bottomrule
\end{tabular}
}

\vspace{1.2em}

\resizebox{\textwidth}{!}{
\rowcolors{3}{gray!15}{white}
\begin{tabular}{ll|c|cccccc}
\toprule
\textbf{Model} & \textbf{Algorithm}
& \multicolumn{1}{c|}{\textbf{New Task Domain}}
& \multicolumn{6}{c}{\textbf{Prior Capabilities}} \\
\cmidrule(lr){3-3} \cmidrule(lr){4-9}
& & Medical
& MMLU & GPQA & AMC & MATH500 & GSM8K & Avg. \\
\midrule

& Base
& \textcolor[HTML]{707070}{0.353}
& \textcolor[HTML]{707070}{0.687}
& \textcolor[HTML]{707070}{0.253}
& \textcolor[HTML]{707070}{0.422}
& \textcolor[HTML]{707070}{0.734}
& \textcolor[HTML]{707070}{0.918}
& \textcolor[HTML]{707070}{0.597} \\

& SFT
& \textcolor[HTML]{98D594}{0.448}
& \textbf{\textcolor[HTML]{005321}{0.614}}
& \textcolor[HTML]{98D594}{0.212}
& \textcolor[HTML]{98D594}{0.193}
& \textcolor[HTML]{98D594}{0.384}
& \textcolor[HTML]{98D594}{0.364}
& \textcolor[HTML]{98D594}{0.353} \\

Qwen2.5-7B-Instruct
& Rewrite SFT
& \textcolor[HTML]{359E53}{0.457}
& \textcolor[HTML]{319A50}{0.571}
& \textcolor[HTML]{17813D}{0.273}
& \textcolor[HTML]{127C39}{0.301}
& \textcolor[HTML]{006D2C}{0.592}
& \textcolor[HTML]{005924}{0.821}
& \textcolor[HTML]{005924}{0.512} \\

& OPSD (on-policy)
& \textbf{\textcolor[HTML]{005321}{0.466}}
& \textcolor[HTML]{53B466}{0.553}
& \textcolor[HTML]{42AB5D}{0.247}
& \textbf{\textcolor[HTML]{005321}{0.337}}
& \textbf{\textcolor[HTML]{005321}{0.625}}
& \textcolor[HTML]{097532}{0.742}
& \textcolor[HTML]{006428}{0.501} \\

& \textbf{Sampling SFT (Ours)}
& \textcolor[HTML]{2E964D}{0.458}
& \textcolor[HTML]{98D594}{0.523}
& \textbf{\textcolor[HTML]{005321}{0.298}}
& \textcolor[HTML]{05712F}{0.313}
& \textcolor[HTML]{005E26}{0.612}
& \textbf{\textcolor[HTML]{005321}{0.833}}
& \textbf{\textcolor[HTML]{005321}{0.516}} \\

\bottomrule
\end{tabular}
}

\vspace{5pt}

\caption{
\textbf{Sampling enables SFT to generalize better while forgetting less.}
Top: Chemistry on Qwen2.5-7B-Instruct.
Middle: Math on Qwen2.5-3B. Bottom: Medical on Qwen2.5-7B-Instruct. We bold the highest score in each column, and for math, we also bold if one of our listed approaches outperforms the other baselines, with a color gradient to emphasize numerical gaps. Across model sizes and evaluation tasks, sampling consistently enables SFT to generalize and retain prior capabilities at least as well as, if not better than, its on-policy posttraining counterparts.}
\label{table:mega_results}

\end{table*}

We display our main results for the Qwen model family in Table \ref{table:mega_results}. For results on Olmo, see Section \ref{sec:analysis} and Appendix \ref{apx:olmo}.

\textbf{Stronger generalization.} Across different pretrained base models and different tasks, we see that projection sampling enables SFT to achieve an exceptionally strong performance when evaluated on new capabilities introduced in finetuning. In chemistry, sampling SFT is able to improve base model performance on chemistry by \textbf{+31.7\%}, exceeding even the improvement yielded by the on-policy OPSD by 
\textbf{+4.20\%}. Simple zero-shot rewrites of the off-policy traces do not perform as strongly in this setting, suggesting the importance of the full MCMC process in Algorithm \ref{alg:samp}. 

Math reasoning vividly illustrates that capability gains transfer to adjacent \textit{out-of-distribution} test domains as well. Whereas vanilla SFT leads to a drop in performance across the board in math benchmarks, sampling SFT yields a \textbf{+18.0\%} boost on the hardest problems in the MATH benchmark, even surpassing the reasoning boosts obtained by GRPO and UFT. This extends to a \textbf{+14.4\%} boost on AMC and a \textbf{+20.3\%} boost on GSM8K, which are on par with the boosts obtained by either RL algorithm. The performance gain on MATH500 is especially notable, providing a \textbf{+33.7\%} boost to the base model and surpassing the next best performing (RL) baseline by \textbf{+26.9\%}.

Not only does the sampling SFT model generalize well, but it also provides a \textit{strong initialization for RL} that leads to even further gains. In the math section of Table \ref{table:mega_results}, we record GRPO initialized with the sampling SFT checkpoint (Sampling SFT-RL) and observe that it is the strongest performing model overall, achieving \textbf{+40.7\%} performance in MATH500 and \textbf{+25.1\%} performance in GSM8K: within reach of the 7B-Instruct model.

With just off-policy expert traces, SFT can fail to generalize well. But composed with our sampling algorithm, SFT can exhibit exceptionally strong generalization, even outperforming RL and OPSD baselines. The fact that this sampling procedure is \textit{custom to the base model distribution} is crucial: Appendix \ref{apx:olmo} details results demonstrating that finetuning, e.g., Qwen on the same base task data but boosted for Olmo severely underperforms all baselines.

\textbf{Retaining prior capabilities.} This strong generalization does not come at the cost of catastrophic forgetting. For chemistry in Table \ref{table:mega_results}, vanilla SFT results in a considerable drop in capabilities. Meanwhile, sampling SFT does not lose any MMLU knowledge and reduces the average loss in prior task accuracy down to just \textbf{-1.10\%}. In fact, our model \textit{forgets the least}, even including OPSD. These results are mirrored in the medical domain, where sampling recovers a \textbf{16.3\%} loss in average prior capabilities with vanilla SFT, and again, forgets the least among all baselines.

\subsection{Analysis}\label{sec:analysis}

We now examine distributional properties of our sampled traces for finetuning as well as those of the resulting model capabilities.

\textbf{Dataset likelihoods.} We can directly observe the effect projection sampling has on expert traces. 

\begin{figure}[h!]
\vspace{-10pt}
  \centering
  \includegraphics[width=\linewidth]{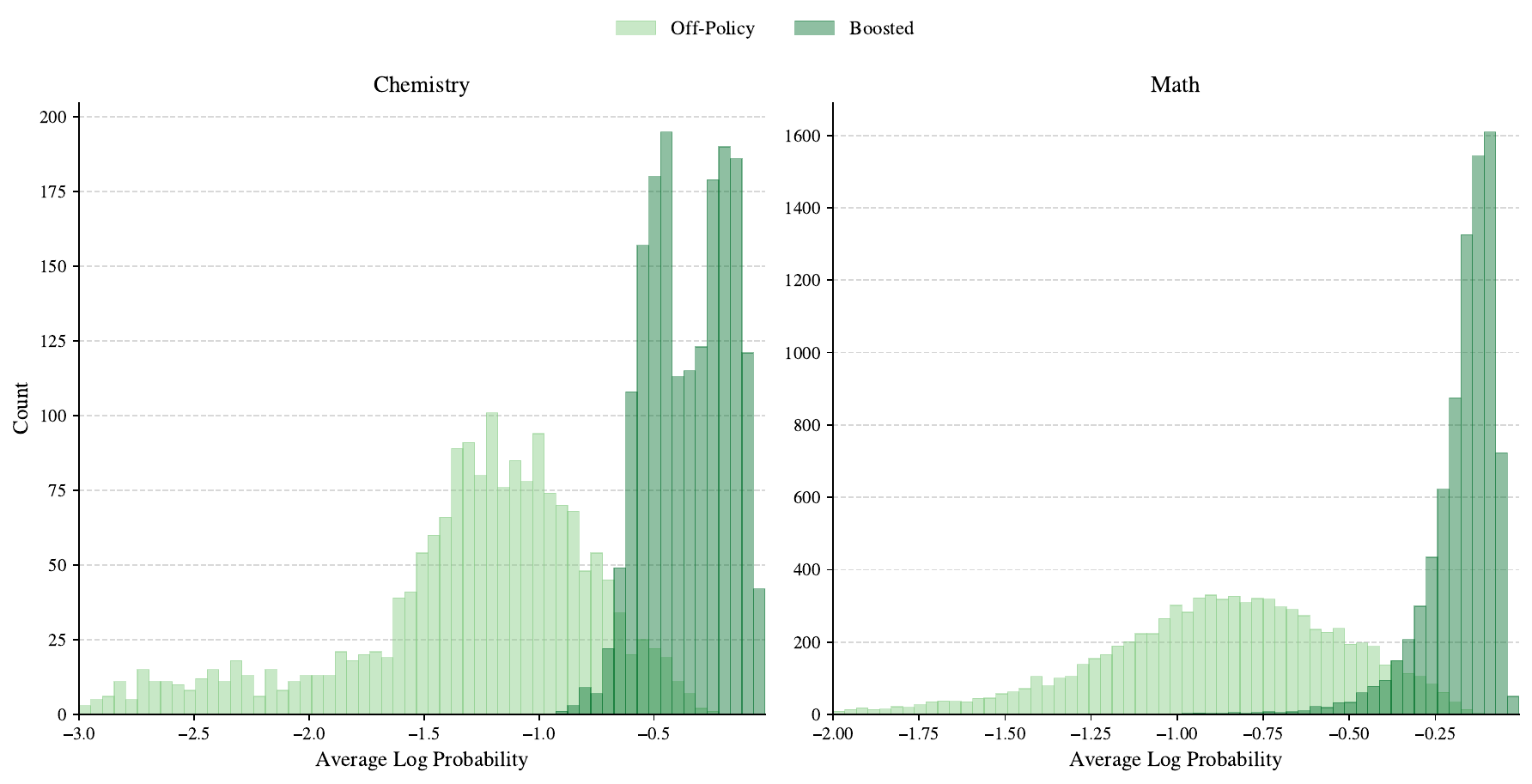}
  \captionsetup{font=small}
  \caption{\textbf{Dataset likelihoods for projection sampling.} We plot likelihoods of the boosted dataset traces for both the chemistry and math task under their corresponding base models (Qwen2.5-7B-Instruct and Qwen2.5-3B).}
  \label{fig:conf}
  \vspace{-8pt}
\end{figure}

Figure \ref{fig:conf} plots histograms of the sequence log-likelihoods (averaged by length) under the base model of both the expert traces and the boosted traces for both science and math reasoning. The on-policy projection is apparent, yielding traces that are much higher-likelihood relative to the base model while preserving correctness. 

\textbf{Learning beyond sharpening.} In Section \ref{sec:intro}, we highlighted the overarching goal of posttraining as \textit{introducing fundamentally new capabilities} that are not present in the base model. Given that projection sampling pushes off-policy data to be more in-distribution, can our finetuned models still acquire new behaviors beyond just sharpening existing ones?

\begin{figure}[h!]
    \centering
    \includegraphics[width=0.65\linewidth]{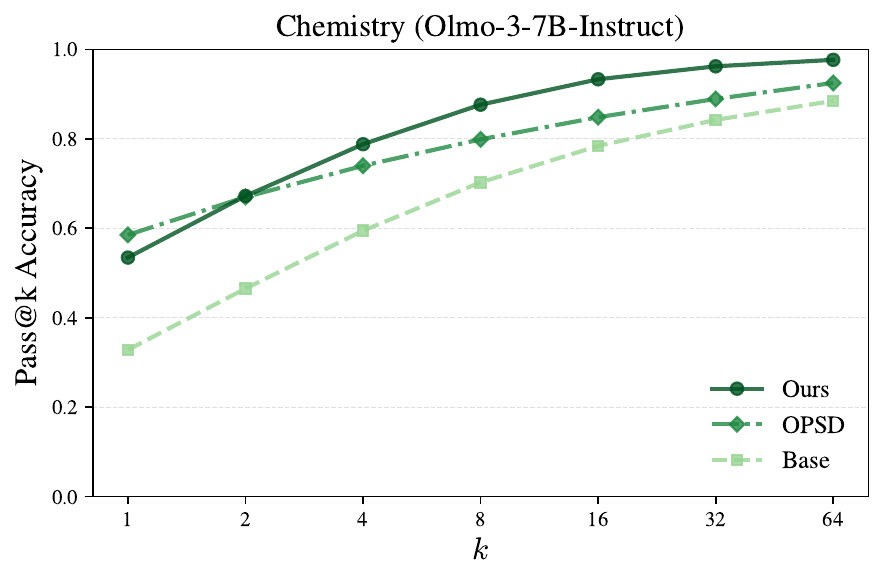}
    \captionsetup{font=small}
    \vspace{-10pt}
    \caption{\textbf{Pass@$k$ performance on Chemistry (Olmo-3-7B-Instruct)}. We plot the pass@$k$ accuracy (correct if at least one of $k$ samples is accurate) of SFT with projection sampling (ours) as well as on-policy learning with privileged information (OPSD) relative to the base model. Our performance curve is strictly better than both OPSD and the base model, and our pass rate at high $k$ exceeds both those of the base model and OPSD.}
    \label{fig:pass}
\end{figure}

We answer in the affirmative. In Figure \ref{fig:pass}, we examine the pass@$k$ accuracy of Olmo-3-7B-Instruct relative to  finetuning with sampling (ours) as well as on-policy distillation with privileged information (OPSD) on the science task. If finetuning with our sampling algorithm was simply sharpening existing capabilities, we would expect our pass@$k$ curve to eventually converge to the base model. However, we actually see large, consistent gaps in the pass@$k$ rate up to very large $k$, demonstrating that our finetuned model has \textit{fundamentally stronger capabilities} than the base model. Moreover, for $k > 2$, we see substantial gaps in our pass@$k$ rate relative to OPSD, indicating that our method enables learning capabilities that OPSD \textit{does not learn}. 

We can also be more granular at the the level of individual evaluation problems. In other words, we can identify ``hard problems'' that the base model absolutely cannot solve (pass@$k$ rate zero for highest $k$) but our finetuned model can start reliably solving: in fact, some evaluation tasks go from zero pass rate in the base model to pass rates that are up to \textbf{67.2\%} or \textbf{53.1\%}.

\textbf{Scaling sampling compute.} Proposition \ref{prop:scale} observes that increasing the number of MCMC steps in projection sampling results in data generation policies $\pi_k$ that are progressively closer to the base model in KL divergence. This frames sampling as a \textit{natural axis for scaling compute}, expending more MCMC steps for more on-policy learning data.

\begin{wrapfigure}{r}{0.6\linewidth}
  \centering
  \includegraphics[width=\linewidth]{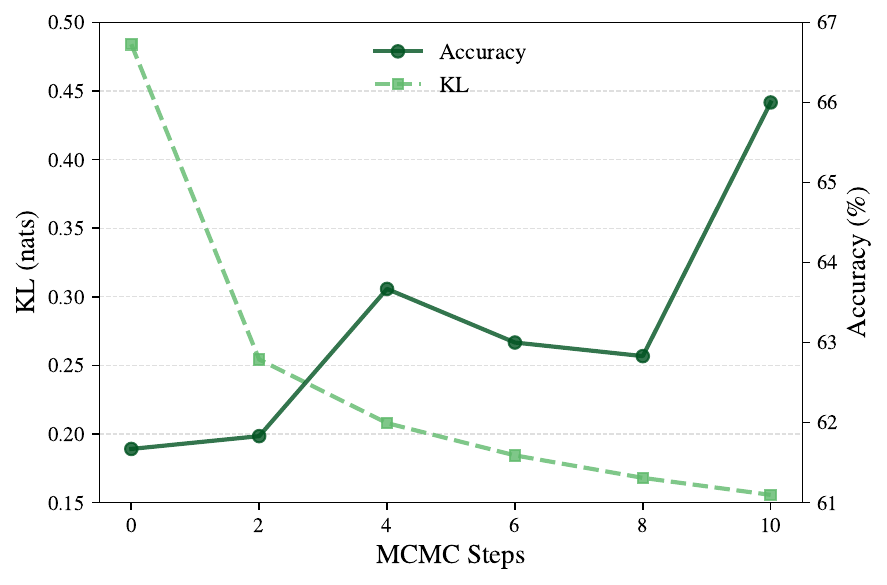}
  \captionsetup{font=small}
  \caption{\textbf{KL divergence vs. accuracy vs. MCMC steps (Qwen2.5-7B-Instruct).}  For $k \in [0, 2, 4, 6, 8, 10]$, we plot both the KL divergence of the boosted data distribution with the base model (Qwen2.5-7B-Instruct) as well as the accuracy of finetuning on the science task. As sampling compute scales, the KL gap decreases while accuracy improves.}
  \label{fig:scale}
  \vspace{-10pt}
\end{wrapfigure}

We can directly observe this scaling empirically in Figure \ref{fig:scale}. For base model Qwen2.5-7B-Instruct on the science task, we apply $k$ MCMC steps to the off-policy training data, for $k \in [0, 2, 4, 6, 8, 10]$. Note that $k = 0$ corresponds to a data rewrite, i.e., asking the base model to simply rewrite off-policy data ``in its own words''. On the left vertical axis, we plot $\mathrm{KL}\!\left(\pi_{k}\,\middle\|\,\pi_{\text{\rm base}}\right)$, where $\pi_k$ is the data generation policy induced by $k$ MCMC steps. We can directly estimate this KL divergence as trajectories are repeatedly generated by a proposal distribution ($\kappa_{\mathcal{C}}( \cdot | \tau)$ in Algorithm \ref{alg:samp}) parametrized by an LLM, from which we can extract next-token log probabilities. On the right vertical axis, we plot the accuracy of the base model finetuned on the boosted data for each $k$. As we increase sampling compute, the boosted data distribution monotonically closes the KL gap with the base model, while the corresponding finetuned models trend upwards in accuracy. Hence, empirically, more MCMC steps lead to more on-policy data, which results in better generalization after finetuning.

\section{Conclusion}

In this work, we introduce a formalism that bridges an apparent disconnect between privileged off-policy information and on-policy training. By carefully boosting the likelihood of off‑policy expert data towards the base model’s distribution, we obtain a more on‑policy equivalent that is far more suitable for learning. Indeed, across science, math reasoning, and open-ended expertise, SFT on boosted data can generalize better and forget less than both vanilla SFT and strong on‑policy baselines, including RL and self‑distillation. The resulting models also exhibit strong distributional performance over multiple samples and are able to learn fundamentally new capabilities that are not present in the base model. 

Beyond just SFT, sampling offers a powerful framework for the posttraining stack, acting as a model-native operator that shapes data for learnability. For example, for on-policy distillation, our MCMC process directly induces a teacher distribution from initial distillation traces that is much closer in KL divergence to the student, facilitating more stable updates. Likewise, conditioned on expert information, our method can simulate on-policy rollouts for RL on hard tasks where the base model is unable to generate signal. By expending sampling-time compute in exchange for stronger, more on-policy trace distributions constructed from privileged information, sampling provides a simple, robust, and scalable pathway for continually expanding LLM capabilities.

\section{Acknowledgements}
A.K. would like to thank the Paul and Daisy Soros Foundation, NDSEG Fellowship, and Kempner
Institute for their support.
S.C. was supported in part by NSF CAREER award CCF-2441635 and the Harvard Dean's Competitive Fund for Promising Scholarship.

\bibliographystyle{plainnat}
\bibliography{main}

@article{shenfeld2025rl,
  title={Rl's razor: Why online reinforcement learning forgets less},
  author={Shenfeld, Idan and Pari, Jyothish and Agrawal, Pulkit},
  journal={arXiv preprint arXiv:2509.04259},
  year={2025}
}

@article{shenfeld2026self,
  title={Self-Distillation Enables Continual Learning},
  author={Shenfeld, Idan and Damani, Mehul and H{\"u}botter, Jonas and Agrawal, Pulkit},
  journal={arXiv preprint arXiv:2601.19897},
  year={2026}
}

@article{cobbe2021training,
  title={Training verifiers to solve math word problems},
  author={Cobbe, Karl and Kosaraju, Vineet and Bavarian, Mohammad and Chen, Mark and Jun, Heewoo and Kaiser, Lukasz and Plappert, Matthias and Tworek, Jerry and Hilton, Jacob and Nakano, Reiichiro and others},
  journal={arXiv preprint arXiv:2110.14168},
  year={2021}
}

@article{hendrycks2020measuring,
  title={Measuring massive multitask language understanding},
  author={Hendrycks, Dan and Burns, Collin and Basart, Steven and Zou, Andy and Mazeika, Mantas and Song, Dawn and Steinhardt, Jacob},
  journal={arXiv preprint arXiv:2009.03300},
  year={2020}
}

@article{feng2024sciknoweval,
  title={Sciknoweval: Evaluating multi-level scientific knowledge of large language models},
  author={Feng, Kehua and Shen, Xinyi and Wang, Weijie and Zhuang, Xiang and Tang, Yuqi and Zhang, Qiang and Ding, Keyan},
  journal={arXiv preprint arXiv:2406.09098},
  year={2024}
}

@article{hastings1970monte,
  title={Monte Carlo sampling methods using Markov chains and their applications},
  author={Hastings, W Keith},
  year={1970},
  publisher={Oxford University Press}
}

@article{xiong2025minimalist,
  title={A minimalist approach to llm reasoning: from rejection sampling to reinforce},
  author={Xiong, Wei and Yao, Jiarui and Xu, Yuhui and Pang, Bo and Wang, Lei and Sahoo, Doyen and Li, Junnan and Jiang, Nan and Zhang, Tong and Xiong, Caiming and others},
  journal={arXiv preprint arXiv:2504.11343},
  year={2025}
}

@article{csiszar1975divergence,
  title={I-divergence geometry of probability distributions and minimization problems},
  author={Csisz{\'a}r, Imre},
  journal={The annals of probability},
  pages={146--158},
  year={1975},
  publisher={JSTOR}
}

@article{guo2025deepseekr1,
  title={Deepseek-r1: Incentivizing reasoning capability in {LLMs} via reinforcement learning},
  author={Guo, Daya and Yang, Dejian and Zhang, Haowei and Song, Junxiao and Zhang, Ruoyu and Xu, Runxin and Zhu, Qihao and Ma, Shirong and Wang, Peiyi and Bi, Xiao and others},
  journal={arXiv preprint arXiv:2501.12948},
  year={2025}
}

@article{hu2025openreasonerzero,
  title={Open-reasoner-zero: An open source approach to scaling up reinforcement learning on the base model},
  author={Hu, Jingcheng and Zhang, Yinmin and Han, Qi and Jiang, Daxin and Zhang, Xiangyu and Shum, Heung-Yeung},
  journal={arXiv preprint arXiv:2503.24290},
  year={2025}
}

@article{hendrycks2021math,
  title={Measuring mathematical problem solving with the {MATH} dataset},
  author={Hendrycks, Dan and Burns, Collin and Kadavath, Saurav and Arora, Akul and Basart, Steven and Tang, Eric and Song, Dawn and Steinhardt, Jacob},
  journal={arXiv preprint arXiv:2103.03874},
  year={2021}
}

@article{li2022alphacode,
  title={Competition-level code generation with {AlphaCode}},
  author={Li, Yujia and Choi, David and Chung, Junyoung and Kushman, Nate and Schrittwieser, Julian and Leblond, R{\'e}mi and Eccles, Tom and Keeling, James and Gimeno, Felix and Dal Lago, Agustin and Hubert, Thomas and Choy, Peter and de Masson d'Autume, Cyprien and Babuschkin, Igor and Chen, Xinyun and Huang, Po-Sen and Welbl, Johannes and Gowal, Sven and Cherepanov, Alexey and Molloy, James and Mankowitz, Daniel and Robson, Esme Sutherland and Kohli, Pushmeet and de Freitas, Nando and Kavukcuoglu, Koray and Vinyals, Oriol},
  journal={arXiv preprint arXiv:2203.07814},
  year={2022}
}

@inproceedings{rein2024gpqa,
  title={{GPQA}: A graduate-level Google-proof Q\&A benchmark},
  author={Rein, David and Hou, Betty Li and Cooper Stickland, Asa and Petty, Jackson and Pang, Richard Yuanzhe and Dirani, Julien and Michael, Julian and Bowman, Samuel R},
  booktitle={First Conference on Language Modeling},
  year={2024}
}

@article{shao2024deepseekmath,
  title={DeepSeek-Math: Advancing mathematical reasoning through step-by-step exploration},
  author={Shao, Zhihong and Wang, Peiyi and Zhu, Qihao and Xu, Runxin and Song, Junxiao and Bi, Xiao and Zhang, Haowei and Zhang, Mingchuan and Li, Y. K. and Wu, Y. and Guo, Daya},
  journal={arXiv preprint arXiv:2404.01140},
  year={2024}
}

@article{metropolis1953equation,
  title={Equation of state calculations by fast computing machines},
  author={Metropolis, Nicholas and Rosenbluth, Arianna W. and Rosenbluth, Marshall N. and Teller, Augusta H. and Teller, Edward},
  journal={Journal of Chemical Physics},
  volume={21},
  number={6},
  pages={1087--1092},
  year={1953},
  publisher={American Institute of Physics},
  doi={10.1063/1.1699114}
}

@article{chen2025retaining,  title={Retaining by Doing: The Role of On-Policy Data in Mitigating Forgetting},  author={Chen, Howard and Razin, Noam and Narasimhan, Karthik and Chen, Danqi},  journal={arXiv preprint arXiv:2510.18874},  year={2025},  url={https://arxiv.org/abs/2510.18874}}

@article{qu2026pope,
  title={POPE: Learning to Reason on Hard Problems via Privileged On-Policy Exploration},
  author={Qu, Yuxiao and Setlur, Amrith and Smith, Virginia and Salakhutdinov, Ruslan and Kumar, Aviral},
  journal={arXiv preprint arXiv:2601.18779},
  year={2026},
  url={https://arxiv.org/abs/2601.18779}
}

@article{ye2026onpolicycontext,
  title={On-Policy Context Distillation for Language Models},
  author={Ye, Tianzhu and Dong, Li and Wu, Xun and Huang, Shaohan and Wei, Furu},
  journal={arXiv preprint arXiv:2602.12275},
  year={2026},
  url={https://arxiv.org/abs/2602.12275}
}

@article{zhao2026selfdistilledreasoner,  title={Self-Distilled Reasoner: On-Policy Self-Distillation for Large Language Models},  author={Zhao, Siyan and Xie, Zhihui and Liu, Mengchen and Huang, Jing and Pang, Guan and Chen, Feiyu and Grover, Aditya},  journal={arXiv preprint arXiv:2601.18734},  year={2026},  url={https://arxiv.org/abs/2601.18734}}

@article{liu2025uft,
  title={UFT: Unifying Supervised and Reinforcement Fine-Tuning},
  author={Liu, Mingyang and Farina, Gabriele and Ozdaglar, Asuman},
  journal={arXiv preprint arXiv:2505.16984},
  year={2025},
  url={https://arxiv.org/abs/2505.16984}
}

@article{chu2025sftmemorizesrlgeneralizes,
  title={SFT Memorizes, RL Generalizes: A Comparative Study of Foundation Model Post-training},
  author={Chu, Tianzhe and Zhai, Yuexiang and Yang, Jihan and Tong, Shengbang and Xie, Saining and Schuurmans, Dale and Le, Quoc V. and Levine, Sergey and Ma, Yi},
  journal={arXiv preprint arXiv:2501.17161},
  year={2025},
  url={https://arxiv.org/abs/2501.17161}
}

@article{li2025blinkofaneyetheory,
  title={Blink of an Eye: A Simple Theory for Feature Localization in Generative Models},
  author={Li, Marvin and Karan, Aayush and Chen, Sitan},
  journal={arXiv preprint arXiv:2502.00921},
  year={2025},
  url={https://arxiv.org/abs/2502.00921}
}

@article{karan2025reguidance,
  title={ReGuidance: A Simple Diffusion Wrapper for Boosting Sample Quality on Hard Inverse Problems},
  author={Karan, Aayush and Shah, Kulin and Chen, Sitan},
  journal={arXiv preprint arXiv:2506.10955},
  year={2025},
  url={https://arxiv.org/abs/2506.10955}
}

@article{gollakota2025predictor,
  title={When Does a Predictor Know Its Own Loss?},
  author={Gollakota, Aravind and Gopalan, Parikshit and Karan, Aayush and Peale, Charlotte and Wieder, Udi},
  journal={arXiv preprint arXiv:2502.20375},
  year={2025},
  url={https://arxiv.org/abs/2502.20375}
}

@article{karan2025reasoning,
  title={Reasoning with sampling: Your base model is smarter than you think},
  author={Karan, Aayush and Du, Yilun},
  journal={arXiv preprint arXiv:2510.14901},
  year={2025},
  url={https://arxiv.org/abs/2510.14901}
}

@article{zhao2024probabilisticinference,
  title={Probabilistic Inference in Language Models via Twisted Sequential Monte Carlo},
  author={Zhao, Stephen and Brekelmans, Rob and Makhzani, Alireza and Grosse, Roger},
  journal={arXiv preprint arXiv:2404.17546},
  year={2024},
  url={https://arxiv.org/abs/2404.17546}
}

@inproceedings{faria2024quest,
  title={QUEST: Quality-Aware Metropolis-Hastings Sampling for Machine Translation},
  author={Faria, Gonçalo R. A. and Agrawal, Sweta and Farinhas, António and Rei, Ricardo and de Souza, José G. C. and Martins, André F. T.},
  booktitle={NeurIPS},
  year={2024},
  pages={},
  url={https://proceedings.neurips.cc/paper_files/paper/2024/file/a221d22ff6a33599142c8299c7ed06bb-Paper-Conference.pdf}
}

@article{lew2023sequential,
  title={Sequential monte carlo steering of large language models using probabilistic programs},
  author={Lew, Alexander K and Zhi-Xuan, Tan and Grand, Gabriel and Mansinghka, Vikash K},
  journal={arXiv preprint arXiv:2306.03081},
  year={2023}
}

@article{ji2026scalable,
  title={Scalable Power Sampling: Unlocking Efficient, Training-Free Reasoning for LLMs via Distribution Sharpening},
  author={Ji, Xiaotong and Tutunov, Rasul and Zimmer, Matthieu and Ammar, Haitham Bou},
  journal={arXiv preprint arXiv:2601.21590},
  year={2026}
}

@article{kirkpatrick2017overcoming,
  title={Overcoming catastrophic forgetting in neural networks},
  author={Kirkpatrick, James and Pascanu, Razvan and Rabinowitz, Neil and Veness, Joel and Desjardins, Guillaume and Rusu, Andrei A. and Milan, Kieran and Quan, John and Ramalho, Tiago and Grabska-Barwinska, Agnieszka and others},
  journal={Proceedings of the National Academy of Sciences},
  volume={114},
  number={13},
  pages={3521--3526},
  year={2017}
}

@article{luo2023empirical,
  title={An empirical study of catastrophic forgetting in large language models during continual fine-tuning},
  author={Luo, Yun and Yang, Zhen and Meng, Fandong and Li, Yafu and Zhou, Jie and Zhang, Yue},
  journal={arXiv preprint arXiv:2308.08747},
  year={2023},
  url={https://arxiv.org/abs/2308.08747}
}

@article{chen2024huatuogpt,
  title={HuatuoGPT-o1, Towards Medical Complex Reasoning with LLMs},
  author={Chen, Junying and Cai, Zhenyang and Ji, Ke and Wang, Xidong and Liu, Wanlong and Wang, Rongsheng and Hou, Jianye and Wang, Benyou},
  journal={arXiv preprint arXiv:2412.18925},
  year={2024}
}

@article{olmo2025olmo3,
  title={Olmo 3},
  author={{Team Olmo} and Ettinger, Allyson and Bertsch, Amanda and Kuehl, Bailey and Graham, David and Heineman, David and Groeneveld, Dirk and Brahman, Faeze and Timbers, Finbarr and Ivison, Hamish and others},
  journal={arXiv preprint arXiv:2512.13961},
  year={2025}
}

@article{karan2024unrolled,
  title={Unrolled Denoising Networks Provably Learn Optimal Bayesian Inference},
  author={Karan, Aayush and Shah, Kulin and Chen, Sitan and Eldar, Yonina C.},
  journal={arXiv preprint arXiv:2409.12947},
  year={2024},
  url={https://arxiv.org/abs/2409.12947}
}

@article{yang2024qwen25,
  title={Qwen2.5 Technical Report},
  author={Yang, An and Yang, Baosong and Zhang, Beichen and Hui, Binyuan and Zheng, Bo and Yu, Bowen and Li, Chengyuan and Liu, Dayiheng and Huang, Fei and Wei, Haoran and others},
  journal={arXiv preprint arXiv:2412.15115},
  year={2024}
}

@article{arzhantsev2026self,
  title={Self-Consistency via Marginal Sharpening},
  author={Arzhantsev, Aleksei and Sakhi, Otmane and Chopin, Nicolas},
  journal={arXiv preprint arXiv:2605.28142},
  year={2026},
  doi={10.48550/arXiv.2605.28142},
  url={https://arxiv.org/abs/2605.28142}
}

@article{wu2025generalization,
  title={On the Generalization of SFT: A Reinforcement Learning Perspective with Reward Rectification},
  author={Wu, Yongliang and Zhou, Yizhou and Zhou, Ziheng and Peng, Yingzhe and Ye, Xinyu and Hu, Xinting and Zhu, Wenbo and Qi, Lu and Yang, Ming-Hsuan and Yang, Xu},
  journal={arXiv preprint arXiv:2508.05629},
  year={2025},
  doi={10.48550/arXiv.2508.05629},
  url={https://arxiv.org/abs/2508.05629}
}

@article{zhu2025anchored,
  title={Anchored Supervised Fine-Tuning},
  author={Zhu, He and Su, Junyou and Lai, Peng and Ma, Ren and Zhang, Wenjia and Yang, Linyi and Chen, Guanhua},
  journal={arXiv preprint arXiv:2509.23753},
  year={2025},
  doi={10.48550/arXiv.2509.23753},
  url={https://arxiv.org/abs/2509.23753}
}

@article{zhou2026reasoning,
  title={Reasoning with Sampling: Cutting at Decision Points},
  author={Zhou, Felix and Mehrotra, Anay and Liu, Quanquan C.},
  journal={arXiv preprint arXiv:2605.30327},
  year={2026},
  url={https://arxiv.org/abs/2605.30327}
}

@article{azizi2026powersmc,
  title={Power-SMC: Low-Latency Sequence-Level Power Sampling for Training-Free LLM Reasoning},
  author={Azizi, Seyedarmin and Potraghloo, Erfan Baghaei and Ahmadi, Minoo and Kundu, Souvik and Pedram, Massoud},
  journal={arXiv preprint arXiv:2602.10273},
  year={2026},
  url={https://arxiv.org/abs/2602.10273}
}


\newpage

\appendix

\section{Examples of Boosted Off-Policy Data}\label{apx:examp}

\begin{center}
\begin{tcolorbox}[
    enhanced,
    width=\linewidth,
    fit to height=11.5cm,
    colback=green!4,
    colframe=green!50!black,
    boxrule=0.5pt,
    arc=2pt,
    title=\small{Chemistry (Olmo-3-7B-Instruct)},
    left=2pt,
    right=2pt,
    top=1pt,
    bottom=0pt
]
{\small
\textit{Question}: Given a chemical equation, please balance the equation and maintain the order of reactants and products as given. Here is an unbalanced chemical equation:
\[
\mathrm{CO}_2 + \mathrm{H}_2\mathrm{O}
=
\mathrm{C}_3\mathrm{H}_6\mathrm{O}_3 + \mathrm{O}_2
\]
Explain your reasoning, thinking step by step, and return the balanced equation on one line. Keep reactants and products in the \textbf{SAME ORDER} as provided.

\vspace{0.8em}
\noindent\rule{\linewidth}{0.4pt}
\vspace{0.2em}

\textit{Original}: To balance the given chemical equation, we need to ensure the number of atoms of each element is equal on both sides of the equation. Starting with carbon (C), there is 1 atom of C in CO$_2$ on the reactant side, but 3 atoms in C$_3$H$_6$O$_3$ on the product side. To balance the carbon atoms, we need 3 CO$_2$ molecules. Next, balance the hydrogen (H) atoms. There are 2 H atoms in each H$_2$O molecule and 6 H atoms in C$_3$H$_6$O$_3$. Therefore, we need 3 H$_2$O molecules. Now, balance the oxygen (O) atoms. We have a total of $3 \times 2 = 6$ O from CO$_2$ and $3 \times 1 = 3$ O from H$_2$O, amounting to 9 O atoms on the reactant side. On the product side, C$_3$H$_6$O$_3$ contributes 3 O atoms, leaving 6 O atoms needed for O$_2$. Since one O$_2$ molecule contains 2 O atoms, 3 O$_2$ molecules are required. The balanced equation is:
\[
3\,\mathrm{CO}_2 + 3\,\mathrm{H}_2\mathrm{O}
\rightarrow
\mathrm{C}_3\mathrm{H}_6\mathrm{O}_3 + 3\,\mathrm{O}_2.
\]

\vspace{1.8em}

\textit{Boosted}: Looking at the given equation, CO$_2$ + H$_2$O $\rightarrow$ C$_3$H$_6$O$_3$ + O$_2$, the first step is to identify the number of atoms of each element on both sides. For carbon, there is 1 C per CO$_2$ molecule on the left, but 3 C atoms in C$_3$H$_6$O$_3$ on the right, so we must place a coefficient of 3 in front of CO$_2$ to balance carbon. This gives us 3 CO$_2$ on the left. Next, we check hydrogen: the right side has 6 H atoms in C$_3$H$_6$O$_3$, while each H$_2$O provides 2 H atoms. So we need 3 H$_2$O molecules (since $3 \times 2 = 6$ H). Now, for oxygen: on the left, 3 CO$_2$ gives 6 O atoms and 3 H$_2$O gives 3 O atoms, totaling 9 O atoms. On the right, C$_3$H$_6$O$_3$ contains 3 O atoms, so the remaining 6 O atoms must come from O$_2$ (since each O$_2$ molecule has 2 O atoms, we need 3 O$_2$ molecules to provide 6 O atoms). Thus, the balanced equation is:
\[
3\,\mathrm{CO}_2 + 3\,\mathrm{H}_2\mathrm{O}
\rightarrow
\mathrm{C}_3\mathrm{H}_6\mathrm{O}_3 + 3\,\mathrm{O}_2.
\]
}
\end{tcolorbox}
\end{center}

\begin{center}
\begin{tcolorbox}[
    enhanced,
    width=\linewidth,
    fit to height=8cm,
    colback=green!4,
    colframe=green!50!black,
    boxrule=0.5pt,
    arc=2pt,
    title=\small{Math (Qwen2.5-3B)},
    left=2pt,
    right=2pt,
    top=1pt,
    bottom=0pt
]
{\small
\textit{Question}: Rationalize the denominator of
$\frac{1+\sqrt{3}}{1-\sqrt{3}}$.
When you write your answer in the form $A+B\sqrt{C}$, where $A$, $B$, and $C$ are integers, what is $ABC$?
Explain your reasoning, thinking step by step, and return your final answer within \texttt{\textbackslash boxed\{\}}.

\vspace{0.8em}
\noindent\rule{\linewidth}{0.4pt}
\vspace{0.2em}

\textit{Original}: When we multiply both the numerator and denominator by the conjugate of the denominator, we get
$\frac{1+\sqrt{3}}{1-\sqrt{3}}
=
\frac{(1+\sqrt{3})(1+\sqrt{3})}{(1-\sqrt{3})(1+\sqrt{3})}$.
Simplifying, we obtain
$\frac{1+2\sqrt{3}+3}{1-3}
=
\frac{4+2\sqrt{3}}{-2}
=
-2-\sqrt{3}$.
Thus $A=-2$, $B=-1$, $C=3$, and $ABC=\boxed{6}$.

\vspace{1.8em}

\textit{Boosted}: To rationalize the denominator of
$\frac{1+\sqrt{3}}{1-\sqrt{3}}$,
we need to multiply both the numerator and denominator by the conjugate of the denominator, which is $1+\sqrt{3}$.

\[
\frac{1+\sqrt{3}}{1-\sqrt{3}}
=
\frac{(1+\sqrt{3})(1+\sqrt{3})}
     {(1-\sqrt{3})(1+\sqrt{3})}
=
\frac{1+2\sqrt{3}+3}{1-3}
=
\frac{4+2\sqrt{3}}{-2}
=
-2-\sqrt{3}
\]

Thus, in the form $A+B\sqrt{C}$, we have $A=-2$, $B=-1$, and $C=3$.
Therefore,
$ABC=(-2)(-1)(3)=\boxed{6}$.
}
\end{tcolorbox}
\end{center}

\begin{center}
\begin{tcolorbox}[
    enhanced,
    width=\linewidth,
    fit to height=8.5cm,
    colback=green!4,
    colframe=green!50!black,
    boxrule=0.5pt,
    arc=2pt,
    title=\small{Medical (Qwen2.5-7B-Instruct)},
    left=2pt,
    right=2pt,
    top=1pt,
    bottom=0pt
]
{\small
\textit{Question}: Given the symptoms of sudden weakness in the left arm and leg, recent long-distance travel, and the presence of swollen and tender right lower leg, what specific cardiac abnormality is most likely to be found upon further evaluation that could explain these findings?

\noindent\rule{\linewidth}{0.4pt}
\vspace{0em}

\textit{Original}: The specific cardiac abnormality most likely to be found in this scenario is a patent foramen ovale (PFO). This condition could allow a blood clot from the venous system, such as one from a deep vein thrombosis in the leg, to bypass the lungs and pass directly into the arterial circulation. This can occur when the clot moves from the right atrium to the left atrium through the PFO. Once in the arterial system, the clot can travel to the brain, potentially causing an embolic stroke, which would explain the sudden weakness in the left arm and leg. The connection between the recent travel, which increases the risk of deep vein thrombosis, and the neurological symptoms suggests the presence of a PFO facilitating a paradoxical embolism.

\vspace{1em}

\textit{Boosted}: Given the patient's history of recent long-distance travel, there is an increased risk of developing a deep vein thrombosis (DVT) due to prolonged immobility during the flight. This DVT, if it occurs in the right lower leg, can lead to swelling and tenderness in that area. Additionally, the presence of a patent foramen ovale (PFO) allows for a potential mechanism by which a clot from the DVT might bypass the pulmonary circulation and travel to the left side of the heart. From there, the clot could enter the systemic circulation and potentially cause an embolic event in the brain, leading to the sudden weakness in the left arm and leg. Therefore, upon further evaluation, the most likely finding would be:

A patent foramen ovale (PFO), which facilitates the passage of a clot from the right side of the heart to the left side, allowing it to reach the cerebral circulation and cause a stroke-like presentation.
}
\end{tcolorbox}
\end{center}

\begin{center}
\begin{tcolorbox}[
    enhanced,
    width=\linewidth,
    fit to height=5cm,
    colback=green!4,
    colframe=green!50!black,
    boxrule=0.5pt,
    arc=2pt,
    title=\small{Math (Qwen2.5-3B)},
    left=2pt,
    right=2pt,
    top=1pt,
    bottom=0pt
]
{\small
\textit{Question}: You are an expert mathematician. Can you solve the following math problem? Suppose that $h(x)=f^{-1}(x)$. If $h(2)=10$, $h(10)=1$ and $h(1)=2$, what is $f(f(10))$? Explain your reasoning, thinking step by step, and return your final answer within \texttt{\textbackslash boxed\{\}}.

\vspace{0.1em}
\noindent\rule{\linewidth}{0.4pt}
\vspace{0.1em}

\textit{Original}: Since $f$ and $h$ are inverse functions and $h(2) = 10$, $f(10) = 2$, so $f(f(10)) = f(2)$.  And since $h(1) = 2$, $f(2) = \boxed{1}$.

\vspace{1em}

\textit{Boosted}: To find $f(f(10))$, we can use the information given about the function $h(x) = f^{-1}(x)$. Since $h(2) = 10$, it means that $f(10) = 2$. Similarly, since $h(1) = 2$, it implies that $f(2) = 1$. Therefore, $f(f(10)) = f(2) = \boxed{1}$.}
\end{tcolorbox}
\end{center}

\begin{center}
\begin{tcolorbox}[
    enhanced,
    width=\linewidth,
    fit to height=8.5cm,
    colback=green!4,
    colframe=green!50!black,
    boxrule=0.5pt,
    arc=2pt,
    title=\small{Medical (Qwen2.5-7B-Instruct)},
    left=2pt,
    right=2pt,
    top=1pt,
    bottom=0pt
]
{\small
\textit{Question}: A 4-year-old boy is brought to the emergency department with a morbilliform rash, fever, cracked lips, cervical lymphadenopathy, and desquamation of the perianal region, suggesting a condition that can lead to serious complications. Which anatomical structure is most critical to evaluate for potential complications in this clinical scenario?

\noindent\rule{\linewidth}{0.4pt}
\vspace{0em}

\textit{Original}: In this clinical scenario, the symptoms presented by the 4-year-old boy suggest Kawasaki disease, a condition especially relevant for its potential complications involving the heart. Most critically, the anatomical structures to evaluate in this case are the coronary arteries. This is because one of the severe complications of Kawasaki disease is the development of coronary artery aneurysms, which can lead to significant cardiac issues. Regular monitoring through echocardiography is often employed to assess the condition of the coronary arteries and ensure timely intervention if needed.

\vspace{1.3em}

\textit{Boosted}: Based on the clinical presentation, the patient's symptoms strongly suggest Kawasaki disease. Given the symptoms such as a morbilliform rash, fever, cracked lips, cervical lymphadenopathy, and desquamation of the perianal region, it is crucial to focus on evaluating the cardiovascular system for potential complications. The coronary arteries are the most critical anatomical structures to assess in this case. This is due to the high risk of developing coronary artery aneurysms, which can occur in up to 25\% of untreated cases. These aneurysms can lead to serious cardiac issues such as myocardial infarction or sudden cardiac death. Therefore, regular echocardiographic monitoring is essential to detect any abnormalities early and provide timely intervention if necessary. In summary, the coronary arteries should be the primary focus when assessing this patient to prevent life-threatening complications associated with Kawasaki disease.
}
\end{tcolorbox}
\end{center}

\newpage

\begin{center}
\begin{tcolorbox}[
    enhanced,
    width=\linewidth,
    fit to height=20cm,
    colback=green!4,
    colframe=green!50!black,
    boxrule=0.5pt,
    arc=2pt,
    title=\small{Chemistry (Qwen2.5-7B-Instruct)},
    left=2pt,
    right=2pt,
    top=1pt,
    bottom=0pt
]
{\small
\textit{Question}: Given a question and four options, think step by step, and then return the correct letter: A, B, C, or D. The last line of your response should be of the following format: $\boxed{{\text{\rm{LETTER}}}}$, where LETTER is one of ABCD (ex. $\boxed{{A}}$).

\bigskip

What is the molar weight (g/mol) of the molecule with the the IUPAC name `(2R,3R,4S,5S)-oxane-2,3,4,5-tetrol'? 

{Options:}

\begin{itemize}
    \item[A)] 148.160
    \item[B)] 150.130
    \item[C)] 151.140
    \item[D)] 149.150
\end{itemize}

Explain your reasoning, thinking step by step, and return your final answer within \texttt{\textbackslash boxed\{\}}.

\vspace{0.8em}
\noindent\rule{\linewidth}{0.4pt}
\vspace{0.2em}

\textit{Original}: To find the molar weight of the molecule '(2R,3R,4S,5S)-oxane-2,3,4,5-tetrol', we first need to determine its molecular formula. The name suggests that it is a tetrahydroxy derivative of oxane, which is a cyclic ether with six atoms (similar in structure to pyran). Since oxane can be considered as tetrahydropyran (with the chemical formula C5H10O), the molecule `(2R,3R,4S,5S)-oxane-2,3,4,5-tetrol' would have four hydroxyl groups (-OH) attached to it. Therefore, the molecular formula would be C5H10O5:

\begin{itemize}
    \item Carbon (C): 5 atoms × 12.01 g/mol = 60.05 g/mol
    \item Hydrogen (H): 10 atoms × 1.008 g/mol = 10.08 g/mol
    \item Oxygen (O): 5 atoms × 16.00 g/mol = 80.00 g/mol
\end{itemize}

Adding these together gives us the molar weight:

\[ 
60.05 \, \text{g/mol} + 10.08 \, \text{g/mol} + 80.00 \, \text{g/mol} = 150.13 \, \text{g/mol}
\]

Therefore, the molar weight of the molecule is 150.130 g/mol. $\boxed{{B}}$

\vspace{1.8em}

\textit{Boosted}: Starting with the IUPAC name `(2R,3R,4S,5S)-oxane-2,3,4,5-tetrol', we can deduce the structure of the molecule. Oxane, also known as tetrahydropyran, has the structure of a five-membered ring containing one oxygen atom. The term ``tetrol'' indicates that this molecule has four hydroxyl (-OH) groups attached to the ring. Therefore, the molecular formula of this compound will be C5H10O5.

Let's break down the calculation of the molar weight step by step:

1. **Carbon (C)**: There are 5 carbon atoms in the molecule. The atomic weight of carbon is approximately 12.01 g/mol.
   \[ 5 \times 12.01 \, \text{g/mol} = 60.05 \, \text{g/mol} \]

2. **Hydrogen (H)**: There are 10 hydrogen atoms in the molecule. The atomic weight of hydrogen is approximately 1.008 g/mol.
   \[10 \times 1.008 \, \text{g/mol} = 10.08 \, \text{g/mol}\]

3. **Oxygen (O)**: There are 5 oxygen atoms in the molecule. The atomic weight of oxygen is approximately 16.00 g/mol.
   \[5 \times 16.00 \, \text{g/mol} = 80.00 \, \text{g/mol}\]

Adding these together gives us the molar weight of the molecule:
\[60.05 \, \text{g/mol} + 10.08 \, \text{g/mol} + 80.00 \, \text{g/mol} = 150.13 \, \text{g/mol}\]

Therefore, the molar weight of the molecule is 150.130 g/mol. $\boxed{{B}}$
}
\end{tcolorbox}
\end{center}

\section{Additional Results}\label{apx:olmo}
We include results on a non-Qwen-family model to demonstrate that our method applies to a variety of pretrained model families. Namely, we provide results for Olmo-3-7B-Instruct on the chemistry task.

\begin{table*}[h!]
\centering

\resizebox{\textwidth}{!}{
\rowcolors{3}{gray!15}{white}
\begin{tabular}{ll|c|cccccc}
\toprule
\textbf{Model} & \textbf{Algorithm}
& \multicolumn{1}{c|}{\textbf{New Task Domain}}
& \multicolumn{6}{c}{\textbf{Prior Capabilities}} \\
\cmidrule(lr){3-3} \cmidrule(lr){4-9}
& & Chemistry
& MMLU & GPQA & AMC & MATH500 & GSM8K & Avg. \\
\midrule

& Base
& \textcolor[HTML]{707070}{0.328}
& \textcolor[HTML]{707070}{0.562}
& \textcolor[HTML]{707070}{0.419}
& \textcolor[HTML]{707070}{0.361}
& \textcolor[HTML]{707070}{0.738}
& \textcolor[HTML]{707070}{0.925}
& \textcolor[HTML]{707070}{0.601} \\

& SFT
& \textcolor[HTML]{81CA81}{0.567}
& \textcolor[HTML]{98D594}{0.568}
& \textcolor[HTML]{98D594}{0.338}
& \textcolor[HTML]{1E8741}{0.422}
& \textcolor[HTML]{98D594}{0.722}
& \textcolor[HTML]{98D594}{0.902}
& \textcolor[HTML]{98D594}{0.590} \\

Olmo-3-7B-Instruct
& Rewrite SFT
& \textcolor[HTML]{98D594}{0.563}
& \textcolor[HTML]{5DB96B}{0.570}
& \textcolor[HTML]{40AA5D}{0.359}
& \textcolor[HTML]{359E53}{0.410}
& \textbf{\textcolor[HTML]{005321}{0.764}}
& \textcolor[HTML]{137D39}{0.916}
& \textcolor[HTML]{329B51}{0.604} \\

& OPSD (on-policy)
& \textbf{\textcolor[HTML]{005321}{0.597}}
& \textcolor[HTML]{17813D}{0.573}
& \textcolor[HTML]{097532}{0.379}
& \textcolor[HTML]{98D594}{0.374}
& \textcolor[HTML]{107A37}{0.754}
& \textbf{\textcolor[HTML]{005321}{0.921}}
& \textcolor[HTML]{4AAF61}{0.600} \\

& \textbf{Sampling SFT (Ours)}
& \textcolor[HTML]{29914A}{0.583}
& \textbf{\textcolor[HTML]{005321}{0.575}}
& \textbf{\textcolor[HTML]{005321}{0.389}}
& \textbf{\textcolor[HTML]{005321}{0.446}}
& \textcolor[HTML]{005C25}{0.762}
& \textcolor[HTML]{137D39}{0.916}
& \textbf{\textcolor[HTML]{005321}{0.617}} \\

\bottomrule
\end{tabular}
}

\vspace{5pt}
\caption{
\textbf{Results on Chemistry using Olmo-3-7B-Instruct. }Chemistry measures
generalization to the new task domain, while MMLU, GPQA, AMC, MATH500,
and GSM8K measure retention of prior capabilities. We report the average
performance across prior-capability evaluations in the final column.
}
\label{table:olmo_chemistry_results}

\end{table*}

We also include two more comparative experiments for the Chemistry task for Qwen2.5-7B-Instruct to stress that the alignment between data distribution and learning distribution is crucial. For the first, we take the boosted dataset relative to Olmo-3-7B and finetune Qwen2.5-7B-Instruct on that. We find the accuracy on Chemistry drops to \textbf{57.33\%}, which is significantly worse than the other baselines (\textbf{61.8\%} for vanilla SFT). For the second experiment, we train Qwen2.5-7B-Instruct on a 50/50 mix of boosted and original SFT data to see the effect of interpolation between off-and-on policy data. We find accuracy on Chemistry reaches \textbf{62.14\%}, in between standard SFT and SFT on the boosted dataset.

\section{Further Experimental Details}
\subsection{Information-Preservation}

One condition on the proposal distribution in Algorithm \ref{alg:samp} to verify is the preservation validity of our MCMC process on off-policy data. We measure this by reporting the accuracy of the boosted dataset for each task: for chemistry, our  dataset retains \textbf{94.33\%} accuracy with Qwen and \textbf{93.94\%} with Olmo. For math, we retain \textbf{95.33\%} and for medical we retain \textbf{95.86\%}. 

\subsection{Sampling}

Note that a full implementation of Algorithm \ref{alg:samp} can require two forward passes through an LLM per iteration: one for generating a candidate $\mathbf{x'}$ via the proposal $\kappa_{\mathcal{C}}$ and another for estimating the transition probability $\kappa_{\mathcal{C}}\left(\mathbf{x} | \mathbf{x'}\right)$. To reduce inference overhead, we approximate the swapping condition by swapping whenever we generate a higher likelihood candidate than the current. This has the effect of much more aggressively shifting to high-likelihood regions under the base model, but as Figure \ref{fig:scale} demonstrates, these aggressive shifts still monotonically reduce the KL divergence between data generation policy and base distribution. 

\subsection{Proposal and Grading Prompts}\label{apx:prompts}
We provide a sample proposal kernel $\kappa_{\mathcal{C}}$ for the chemistry task below (other tasks follow a similar prompt) as well as the semantic equivalence grading prompt for the medical task that is fed into GPT5-mini.
\begin{promptbox}
You are an expert chemist given a chemistry problem, its solution, and an initial, partial response. Carefully study the solution,
identifying what reasoning or steps are already provided, and then continue the partial response. Ensure your response is logically consistent with the solution and leads to a complete and correct final answer.

\bigskip

Task: <PROBLEM>

\bigskip

This is an example for a solution to the problem: <SOLUTION>

\bigskip

Here is a partial response: <RESPONSE>

\bigskip

Starting with the partial response, continue the response in your own words, including the thinking process. Ensure the final answer exactly matches that of the provided solution.

\end{promptbox}

\begin{promptbox}
You are an expert medical evaluator assessing whether a model’s response correctly answers a medical question. Your task is to compare the model’s response to the reference answer and determine if the model’s response is:
1. CORRECT: The response contains the key medical information from the reference answer, even if phrased differently or includes additional correct medical details. 2. INCORRECT: The response is medically wrong, misses the main point, or provides incorrect medical information. Focus on medical accuracy and completeness, not on writing style or
verbosity. 

\bigskip

[Medical Question]: <QUESTION>

[Reference Answer]: <ANSWER>

[Model Response]: <RESPONSE>

\bigskip

Evaluate the model’s response. Output ONLY one of: "CORRECT" or "INCORRECT".
\end{promptbox}

\end{document}